# Indicating Robot Vision Capabilities with Augmented Reality

Hong Wang[1] · Ridhima Phatak[1] · James Ocampo[1] · Zhao Han[1]

**Abstract**

Research indicates that humans can mistakenly assume that robots and humans have the same field of view, possessing an inaccurate mental model of robots. This misperception may lead to failures during human-robot collaboration tasks where robots might be asked to complete impossible tasks about out-of-view objects. The issue is more severe when robots do not have a chance to scan the scene to update their world model while focusing on assigned tasks. To help align humans' mental models of robots' vision capabilities, we propose four field-of-view indicators in augmented reality that reveal the robot's actual horizontal vision limitations to human collaborators and conducted a human-subjects experiment ($N = 41$) to evaluate them in a collaborative assembly task regarding accuracy, confidence, task efficiency, and workload. These indicators span a spectrum of positions: two at robot's eye and head space—deepening eye socket and adding blocks to two sides of the eyes (i.e., egocentric), and two anchoring in the robot's task space—adding extended blocks from the sides of eyes to the table and placing blocks directly on the tables (i.e., allocentric). Results showed that, when placed directly in the task space, the allocentric indicator yields the highest accuracy, although with a delay in interpreting the robot's field of view. When placed at the robot's eyes, the egocentric indicator of deeper eye sockets, possible for physical alteration, also increased accuracy. In all indicators, participants' confidence was high while cognitive load remained low. Finally, we contribute six guidelines for practitioners to apply our augmented reality indicators or physical alterations to align humans' mental models with robots' vision capabilities.

## 1 Introduction

Mental models are structured knowledge systems that enable people to engage with their surroundings [60]. They influence how people perceive problems and decision-making [24], and show how individuals interact within complex systems, such as technological or natural environments [35]. In a team environment, a shared mental model improves team performance when team members have a mutual understanding of each other's roles and the collaborative task [25]. This is also true in technological environments like human-agent teams [52], applicable to physically embodied agents like robots.

Indeed [36], found that both team- and task-based mental models were positively related to efficient team process and performance. This highlights the importance of shared mental models in shaping effective teamwork. To leverage the shared mental models [17], proposed a cooperative inverse reinforcement learning formulation to ensure that agents' behaviors are aligned with humans' goals. [40] also developed a game-theoretic model of human adaptation in human-robot collaboration. These studies show that shared mental models are beneficial for both human teams and human-robot teams: They enhance coordination, improve performance, and help understand collaborative tasks.

✉ Zhao Han
zhaohan@usf.edu

Hong Wang
hongw@usf.edu

Ridhima Phatak
phatakr@usf.edu

James Ocampo
jamesocampo@usf.edu

1 Bellini College of Artificial Intelligence, Cybersecurity and Computing, University of South Florida, 4202 E. Fowler Avenue, Tampa, FL 33620, USA

However, in human-robot teaming and collaboration scenarios, because robots more or less resemble humans, humans can form an inaccurate mental model of robots' capabilities, leading to mental model misalignment. [14] noticed this problem and proposed an asymmetric interaction model: Unlike symmetric interaction models where roles and capabilities are mirrored between humans and robots, asymmetric interaction models emphasize the distinct strengths and limitations of humans and robots.

One mental model misalignment case related to a robot's vision limitation is the assumption that robots possess the same field of view (FoV) as humans. Although humans have over 180° FoV, a robot's camera typically has less than 60° horizontal FoV (e.g., Pepper's 54.4° [4] and Fetch's 54° [47, 61]). This discrepancy and assumption are problematic. Particularly, our previous work [18] studied how a robot can convey its incapability of handing a cup, both out of reach and out of view, and found that participants assumed human's FoV and demanded an explanation that was not needed with a correct mental model.

Specifically, in the dynamic scenario [18], a robot was completing an organization task in front of a table while a person was busy watching a video on a laptop on the right end of the table. The person became thirsty and wanted the robot to pass a cup that the person left on the left end of the table, asking "Can you pass the cup?" However, the robot did not have a chance to move its head to scan the scene to add the cup to its world model while busy organizing the middle part of the table. Despite the cup being out of the robot's less-than-60° FoV, participants assumed the cup was within the robot's FoV, and expected the robot to successfully hand it to them. In this case, the robot can move its head to scan the scene. However, if it scans the right first, the person will be confused and wonder why it did not look left to take the cup, demanding explanations. This dynamic environment and such misalignment highlight the importance of developing an accurate mental model of the robot's real vision capability, even when the robot could scan the scene to find the cup. If people form a correct mental model before the request, it will lead to fewer explanations and clearer instructions, e.g., asking "the cup on the right" rather than "the cup".

In this paper, we aim to address the FoV discrepancy by FoV indicators, answering *"how would a robot indicate its limited FoV to align humans' mental model of robots' vision capabilities?"* Towards this end, we first explored the design space with a taxonomy from eye/head space (egocentric designs) to task space (allocentric designs), informed by the taxonomy, proposed four indicators, and conducted a human-subjects study to evaluate them. Specifically, we designed and registered four augmented reality (AR) indicators (Fig. 1b) to a Pepper robot and conducted a human-subjects study ($N$=41) to investigate the effects of those designs.

In the study, participants followed four instructions to assemble a partially built airplane model with the help of a robot, which participants requested for objects if they believed the robot could see the objects. This is part of a human-robot collaborative task where the robot may not be able to scan the scene at request time to update its world model, inspired by the dynamic handover scenario from our previous work [18]. To summarize, while the robot can scan the scene, such behavior has three drawbacks: (1) Scanning in the wrong direction will cause confusion and lead to an unnecessary demand for explanations; (2) The robot may not be able to scan the scene to overcome its limited FoV while busy working on its part, e.g., manipulation; (3) The scan adds delays to task completion time.

Among our four designs, the first two can be physical alterations or additions to the robot, and the other two were in AR. AR is of interest for four reasons: (1) Robots' hardware, e.g., eye socket, is hard to modify after fabrication and AR allows overlaying the modification image (see design Eye Sockets in Fig. 1b); (2) AR allows fast prototyping for exploring multiple designs and adaptation to changes in an iterative design process [57]; (3) AR allows situated visualizations [51] in relevant contexts, which are the task environment and the eye area; (4) AR was recently found to be equivalent to their physical counterpart in both objective and subjective metrics after comparing an AR vs physical arm attached to a physical mobile robot in a reference task [19].

## 1.1 Contribution

In summary, our contributions are fourfold:

(1) We proposed a taxonomy and spectrum to categorize field-of-view indicators from egocentric (eye and head space) to allocentric (task space) indicators for conveying robot vision capabilities.
(2) We proposed and implemented four AR FoV indicator designs (two egocentric, one transition-space, one task-space), registered onto a Pepper robot and its workspace.
(3) Through a mixed-design human-subjects study ($N$=41) in a collaborative assembly task, we contribute empirical evidence regarding accuracy, confidence, efficiency, and workload. Specifically, we found all four indicators improved FoV accuracy over the Baseline, with Blocks at Task achieving the highest accuracy and Eye Sockets also relatively accurate; Extended Blocks yielded the shortest completion time, whereas Blocks at Task incurred longer but more accurate interactions. Confidence ratings were generally high and workload was generally low, with no statistically significant differences among all designs.

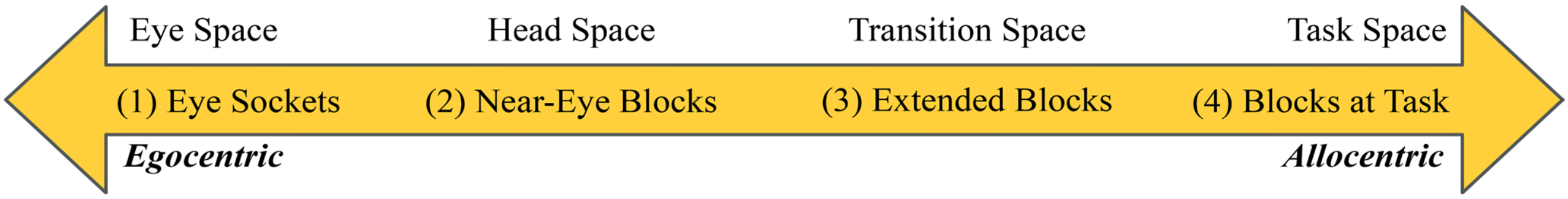

(a) Our design spectrum: from the robot (eyes to head) towards the task environment, and vice versa.

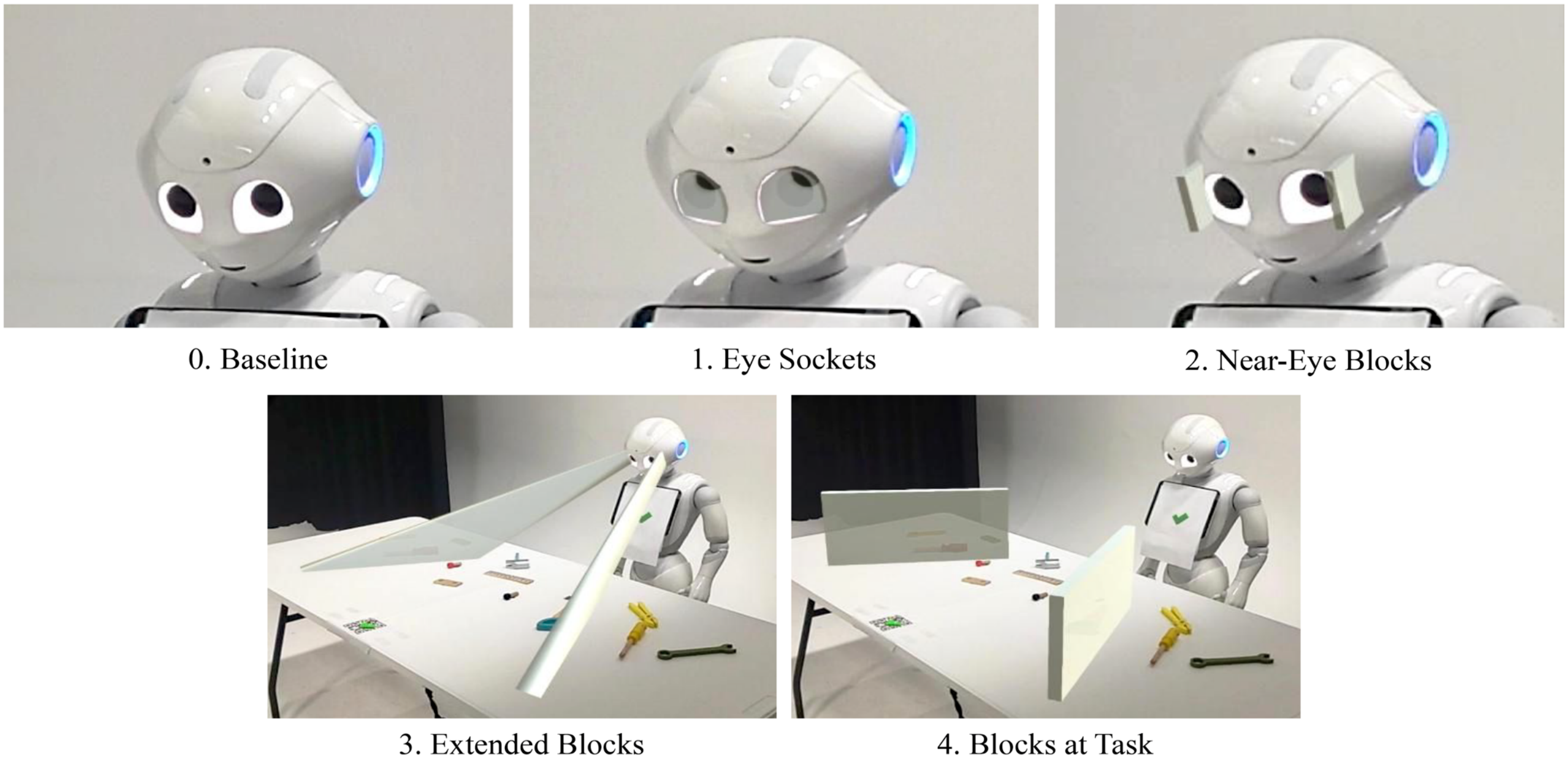

(b) To indicate a robot's vision capability, i.e., the field of view (FoV), we propose four egocentric and allocentric indicators in augmented reality (AR) and evaluated them in a user study with Baseline (no indicators). The design philosophy–from the eyes/head to task space–and descriptions of each design are detailed in Section 3 and 4.

**Fig. 1** AR indicators for communicating a robot's field of view (FoV) across our spatial spectrum. The top panel illustrates the egocentric–allocentric spectrum and the four spatial regions; the bottom panel shows the four AR overlays on Pepper and the workspace, together with the Baseline condition (0) without an indicator

(4) To conclude, we contribute **six design guidelines** for practitioners who wish to apply AR indicators or physical alterations to improve the transparency of robot vision. As a preview, the design guidelines are

- (i) Design Guideline 1: Without other AR indicators, robot designers should design deeper eye sockets to match each camera's FoV
- (ii) Design Guideline 2: If AR situated visualization can be leveraged, robot designers should add FoV indicators at the task space for nearly perfect accuracy.
- (iii) Design Guideline 3: Robot designers should connect AR FoV indicators at the task space to the eyes for efficiency.
- (iv) Design Guideline 4: If Extended Blocks is used alone, robot designers should be aware that wrong guessers might be overconfident.
- (v) Design Guideline 5: Robot designers should rest assured that although the highly accurate FoV indicator at the task space has lower task efficiency, the workload has remained low.
- (vi) Design Guideline 6: For mission-critical collaborative tasks that require accuracy, the allocentric design like Blocks at Task should be used.

## 2 Related Work

### 2.1 AR for Robotics

Robotics researchers have integrated AR in multiple domains. Some examples include AR systems for fault visualization in industrial robots [5], integration in robotic surgical tools [10], and AR-enhanced robotics education for interaction [44]. These studies showed the potential of AR in enhancing HRI. Particularly [5], developed an AR system to support fault visualization in industrial robotic tasks by visualizing the robot's operational status and faults with icons in AR. In the study by [26], AR was integrated into a heterogeneous fleet management system with AR-equipped safety vests for human workers, allowing them to see where the robots are out of sight, e.g., blocked by racks , thereby increasing the feeling of safety in warehouse environments. [10] surveyed the integration of AR/VR with robotic surgical tools, showing that AR overlays increased precision and user comprehension in complex surgeries. For a comprehensive survey, we refer readers to [57].

While these works focused on improving performance, interaction, and understanding of the tasks at hand in various HRI contexts, they did not address the wrong human mental model of a robot's real capabilities like vision. Our work bridges this gap using AR design elements.

### 2.2 AR Design Elements for HRI

We explored how AR design elements enhanced HRI in the past. According to [57], virtual design elements in AR are visualizations that augment robot interactivity, including user-anchored visualizations and robot- or environment-anchored elements. They proposed four virtual design element categories: Virtual entities, virtual alterations, robot status visualizations, and robot comprehension visualizations.

*Virtual entities* add virtual objects, robots, or environments to the user's view. Examples include "visualization robots" that reveal hidden robot poses in teleoperation and digital twins that help predict future actions by superimposing picking poses and robot trajectories [31]. These works show how additional virtual objects can reveal aspects of robot behavior that are otherwise invisible; our FoV indicators similarly add virtual elements so that people know what the robot can and cannot see.

*Virtual alterations* modify a robot's appearance using virtual imagery. For instance [5], used cosmetic alterations to highlight an industrial robot's joint in red to draw attention quickly when a fault occurs (e.g., lifting heavy objects that exceed payload limitation). [56] overlaid arrows and eyes to an aerial robot to signal its navigation intent. [16] and [19] took this concept further by adding virtual body extensions–virtual arms to use gestures to naturally communicate with humans. Inspired by this line of work, our Eye Sockets and Near-Eye Blocks designs alter Pepper's eye appearance to communicate its FoV.

*Robot status visualizations* communicate the robot's internal and external states to users. Internal visualizations display information like battery levels or actuator status directly within the user's view, e.g., next to a stereo video stream [5, 30]. These visual elements help users monitor the robot's condition and identify potential issues such as sensor malfunctions or actuator faults. External state visualizations provide information about a robot's current pose and motion plan, helping maintain situational awareness [9].

To be described in Section 4, our designs fall under "Virtual Alterations – Morphological" (designs 1 and 2) and "Virtual Entities - Environmental" (designs 3 and 4). Yet, we focus on enhancing the comprehension of the robot's FoV.

### 2.3 AR for Robot Comprehension

The most relevant category to our work is the fourth one: *Robot comprehension visualizations*, which convey the robot's beliefs of its environment and tasks. [13] proposed a mobile AR interface to show the regions a robot can physically reach. [49] developed a debugging tool in AR to show the navigation goals to enhance the transparency of mobile robots. In a user study [48], additionally showed that using head-mounted displays to visualize the robot's motion plan, like arm movements, improved task accuracy and speed compared to traditional 2D display methods. For drones [53], explored the design space of visually communicating the directional intentions of drones using AR. Together, these studies demonstrate that AR can effectively reveal a robot's planned motions, goals, and reachable space.

Another line of work focuses on conveying what a robot perceives about the environment to people, e.g., adding external sensor purviews [57]. Entity labels such as part identifiers, operational status, and next action steps can be projected directly into the workspace with projector-based AR to show the robot's planned trajectories and task states, enhancing transparency and coordination [7]. Kobayashi et al. [29] used AR to overlay obstacle representations and decision-making processes of navigation onto the physical environment. [46] showed how robots understand and navigate their environment by aligning visual maps and highlighting key areas or objects. These approaches make the robot's perception and plans visible, but they do not directly address how people understand the *limits* of a robot's FoV in collaborative interaction.

The most relevant work is [21]'s. They developed three teleoperation models to provide visual feedback on robot camera capabilities like real-time visual overlays, interactive interface elements, and enhanced camera feeds. However, their work has focused on non-collocated teleoperation. Our work, while aligning with robot comprehension visualizations in environments, specifically aims to convey robots' vision capabilities in-situ.

## 3 Taxonomy and Spectrum

As robots are physically situated in our physical world, we categorized our designs into four connected areas between the robot and its operating environment: Eye Space, Head Space, Transition Space, and Task Space. It formed a spectrum as shown in Fig. 1b.

*Egocentric* designs focus on the modifications at the robot's eyes, which possess the property of FoV, or near its head. Examples include designs 1 and 2, directly influencing the robot's ability to perceive its surroundings.

Expanding rightwards, *transition space* includes the design that extends from the robot into its operating environment, such as design 3 in Figure 1b. This design bridges the gap between the robot and the environment. As the indicator moves closer to the task setting, we hypothesize that this design will better help people identify the performance effects of FoV. Finally, design 4 in *allocentric* in Fig. 1b is not attached to the robot but rather placed in its working environment. Spectrum in Figure 1a offers a visual breakdown of our indicator designs, emphasizing the continuum from the robot space to the environment space.

## 4 Field-of-View Indicators

Based on the taxonomy, we proposed four indicators. Initially, we had nine designs [58]. However, our pilot studies showed that experiencing four designs took half an hour, and to avoid fatigue effects affecting results, we therefore selected one representative for each space: Eye Sockets (eye space), Near-Eye Blocks (head space), Extended Blocks (transition space), and Blocks at Task (task space). We left the other five designs for future work and are working on implementing and evaluating them. The number prefixes below are the same as in Fig. 1b.

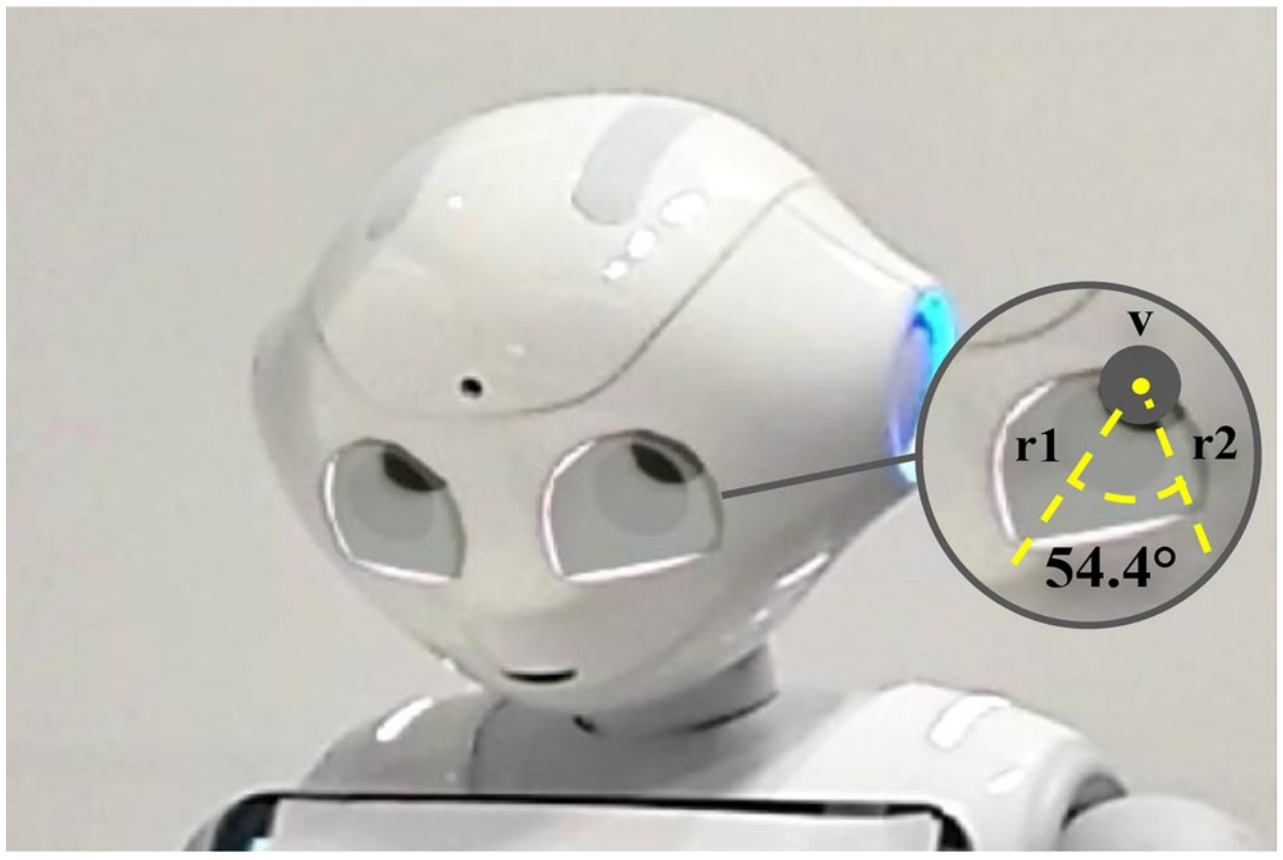

**Fig. 2** Illustration of eyeball depth calculation in the eye socket design to match the robot's FoV

**(1) Eye Socket**: As an egocentric design, we deepen the robot's eye sockets using an AR overlay at the existing eye sockets. It creates a deeper eyeball in the robot's eyes. As the sockets deepen with a deeper eyeball, they appear to physically limit what angle the eyes can see, thus matching the cameras' FoV. This design is possible both physically and in AR, but physical alteration is difficult after fabrication.

For our implementation, we calculated the eye socket depth to match the Pepper robot's 54.4° FoV by calculating the angle $\measuredangle v$ based on the outer dimension of the socket boundaries. Specifically, given the eyeball center as endpoint $v$, we form a ray $r_1$ starting from $v$ and passing the left socket edge, and another ray $r_2$ also starting from $v$ but passing the right socket edge. The eyeball was deepened until the angle $\measuredangle v = 54.4°$ between $r_1$ and $r_2$, as seen in Figure 2.

**(2) Near-Eye Blocks**: We add blocks directly to the sides of the robot's eyes to functionally block those outside of the camera's FoV. This design is possible both physically and in AR.

**(3) Extended Blocks**: To more accurately show the range of the robot's FoV (e.g., which objects the robot cannot see), we connect the blocks from the robot's head (eye sides) to the task environment, so people know exactly how wide the robot can see. Note that this design can only be practically made possible with AR.

**(4) Blocks at Task**: An egocentric or task-centric design is to place the blocks directly in the robot's task environment to show the robot's FoV, e.g., a table. Unlike Extended Blocks, this is in the environment rather than connected to the robot. Note that this design can also only be placed with AR.

## 5 Hypotheses

As the **indicators are increasingly closer to the task space**, towards the right end of the spectrum in Fig. 1b, we believe they will bring task-related and subjective benefits. Thus, we develop the following four hypotheses (H).

**H1**: Participants will have a more accurate mental model of robots' vision capabilities. This will be measured by the percentage of correct guesses of whether objects are within or outside the robot's FoV.

**H2**: Indicators towards the task environment will improve task efficiency because less time will be spent on guessing whether the robot can fulfill the requests or for the robots to ask clarification questions.

**H3**: Participants will be more confident in their guesses. This will be measured by a sevenpoint Likert scale question.

H4: Designs closer to the task environment will require less cognitive effort. This will be measured by the well-established NASA Task Load Index [20, 39].

## 6 Method

To test our hypotheses, we designed a mixed-design human-subjects study with five conditions (Table 1): Baseline (design 0, egocentric), Eye Sockets, Near-Eye Blocks, Extended Blocks, and Blocks at Task. The three egocentric indicators (design {0,1,2}) were tested within subjects using a balanced Latin square. To avoid learning effects for the allocentric designs that explicitly reveal the robot's FoV, each participant experienced only one allocentric indicator (design {3,4}), yielding a between-subjects comparison for these two conditions.

**Table 1** Counterbalanced ordering to control ordering effect: latin square ordering for design {0,1,2}, and, then, fully counterbalanced ordering for design {3,4} because design {3,4} reveals the FoV

| Participant | {0,1,2} Order | {3,4} Order |
|---|---|---|
| 1 | 0, 1, 2 | 3 |
| 2 | 1, 2, 0 | 4 |
| 3 | 2, 0, 1 | 3 |
| 4 | 0, 1, 2 | 4 |
| 5 | 1, 2, 0 | 3 |
| 6 | 2, 0, 1 | 4 |
| … | … | … |

### 6.1 Apparatus and Materials

**Robot Platform**: We used a Pepper robot [42] manufactured by Aldebaran. It is a two-armed, 1.2 m (3.9 ft) tall humanoid robot. Its narrow horizontal FoV is 54.4° [4], commonly seen in other robots like Fetch [47, 61].

**AR Display**: Participants wore a Microsoft HoloLens 2, an optical see-through head-mounted display [37]. It has a 43° ×29° FoV and 2048×1080 resolution per eye. To compensate for the limited FoV, participants were instructed to move along the table to check the design from multiple perspectives.

**Toolkit Set**: A toolkit set [28] was used for an airplane model assembly task. It has six types of tools (2 wrenches, 2 screwdrivers, 1 plier, 1 hammer, 1 saw, and 1 ruler) as shown in Figure 3 part (b) and five types of assembly parts (9 assembly pieces, 3 building blocks, 4 wheels, 6 bolts, and 5 nuts) as shown in Figure 3 part (a).

**Tables and Object Placement**: Two 182 cm × 76 cm tables [38] (Fig. 4) were placed in front of the robot (robot table) and participants (task table). To mimic real-world settings, we randomly clustered 12 objects taken from the toolkit with different object orientations. With tape, we marked the positions of the objects, tables, and the robot to ensure consistency across all conditions throughout the experiment.

### 6.2 Task

As seen in Fig. 5, participants were tasked to follow four instructions to finish a partially assembled airplane using four objects on the robot table (see Fig. 6): A red screw, a red screwdriver, a blue screw, and a yellow screwdriver.

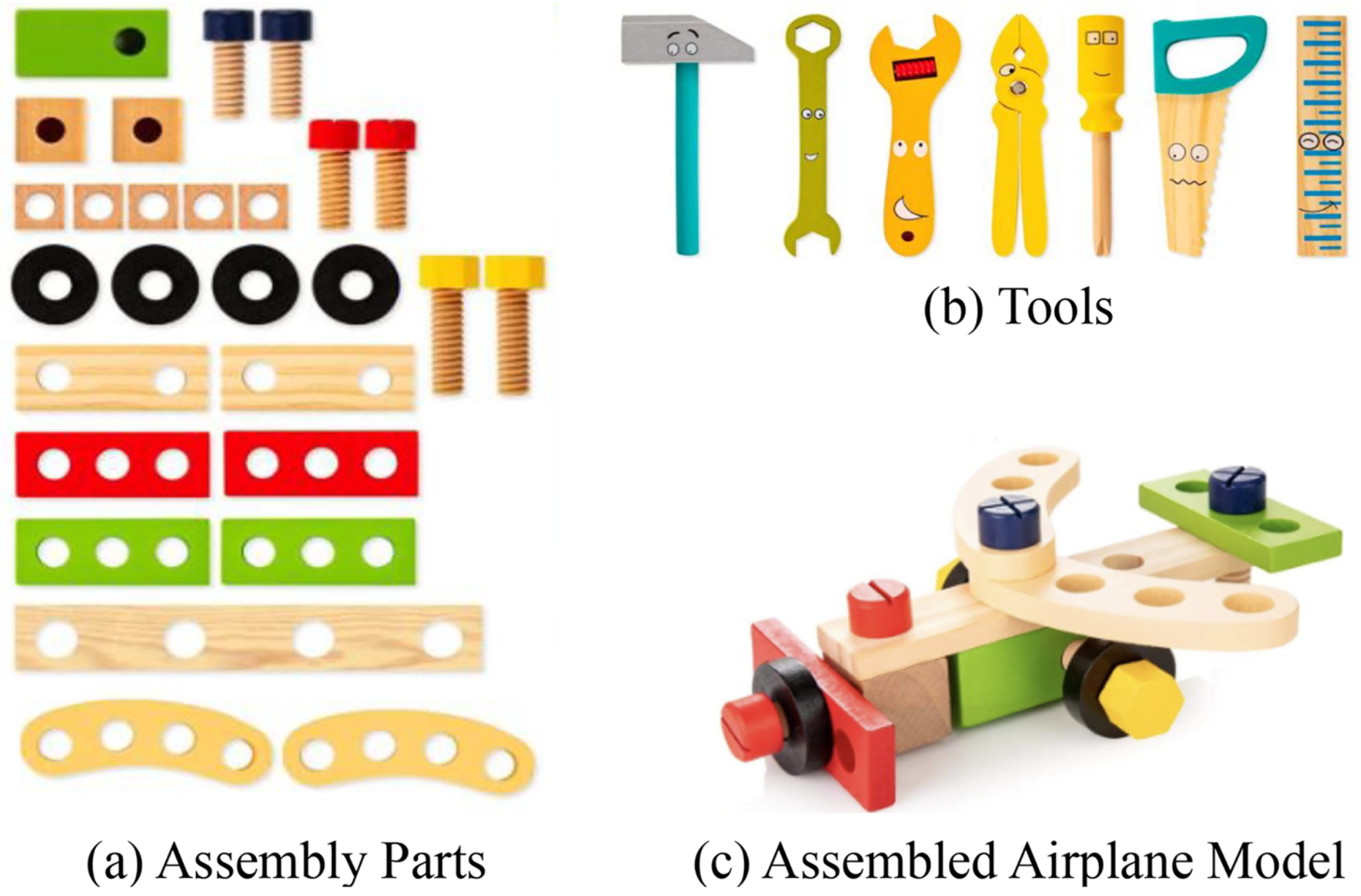

**Fig. 3** The toolkit used in our collaborative task. (product photo [28] used under fair use.)

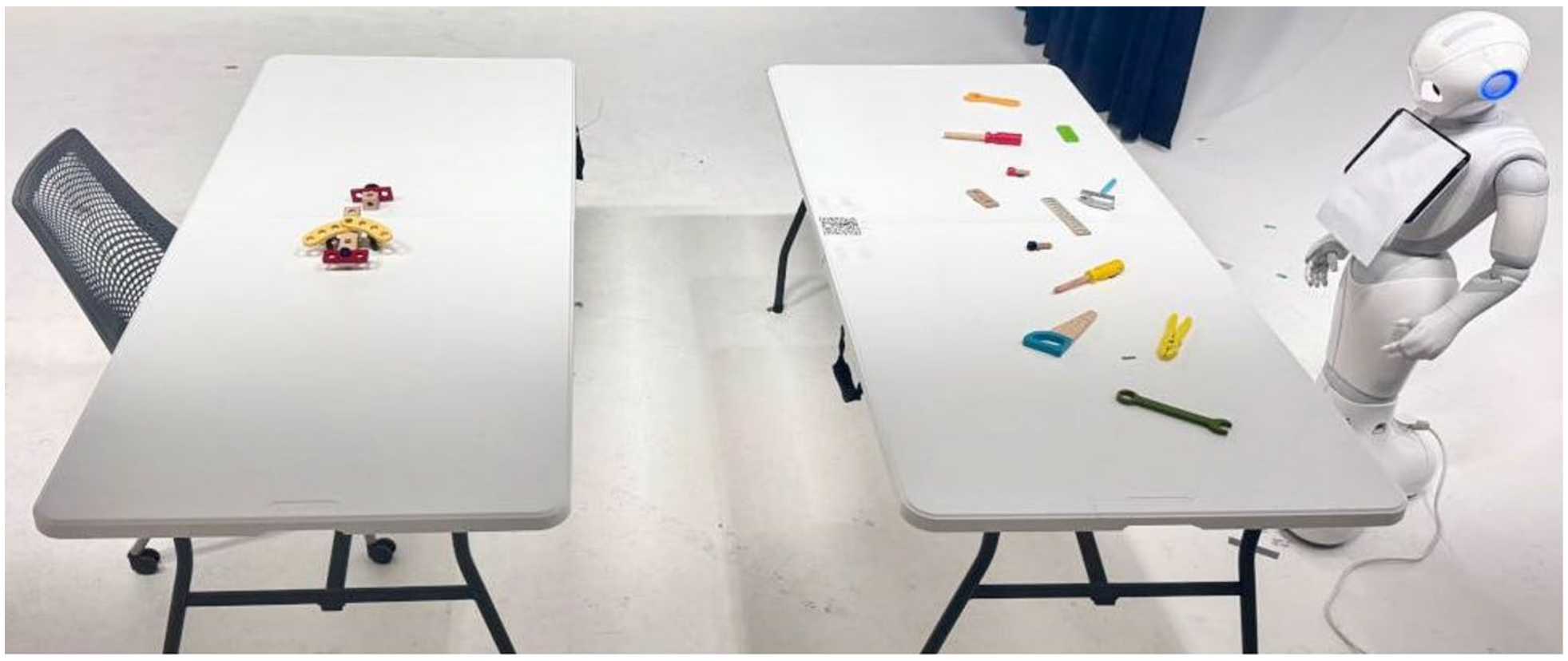

**Fig. 4** Experiment setup. *left task Table*: two pre-assembled parts for participants to start building the airplane model. Participants sat approximately 3.3 meters away so they could see the indicators in full. *right robot Table*: objects within the robot's reach and needed to finish assembly

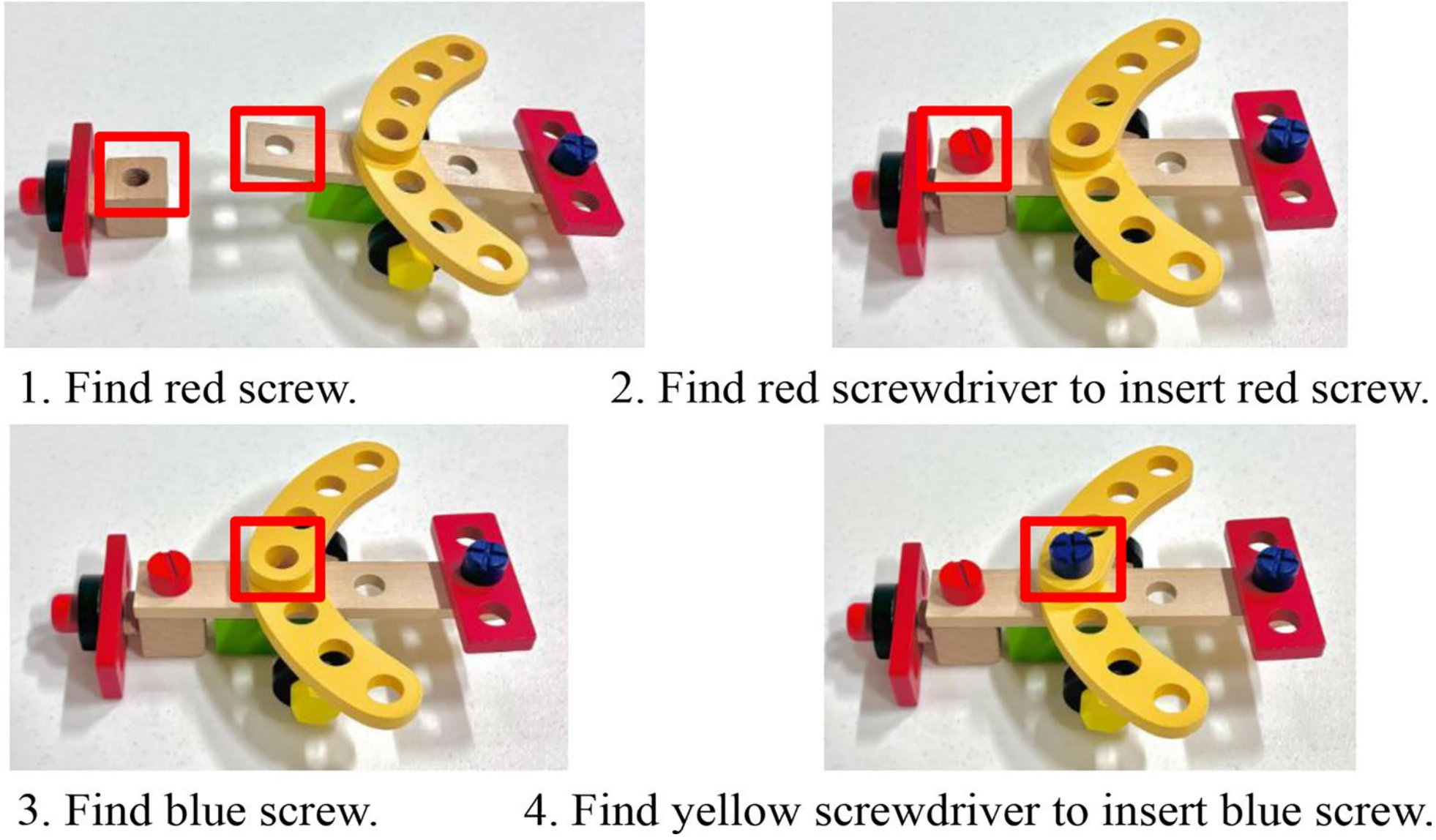

**Fig. 5** Four assembly steps to build the airplane model

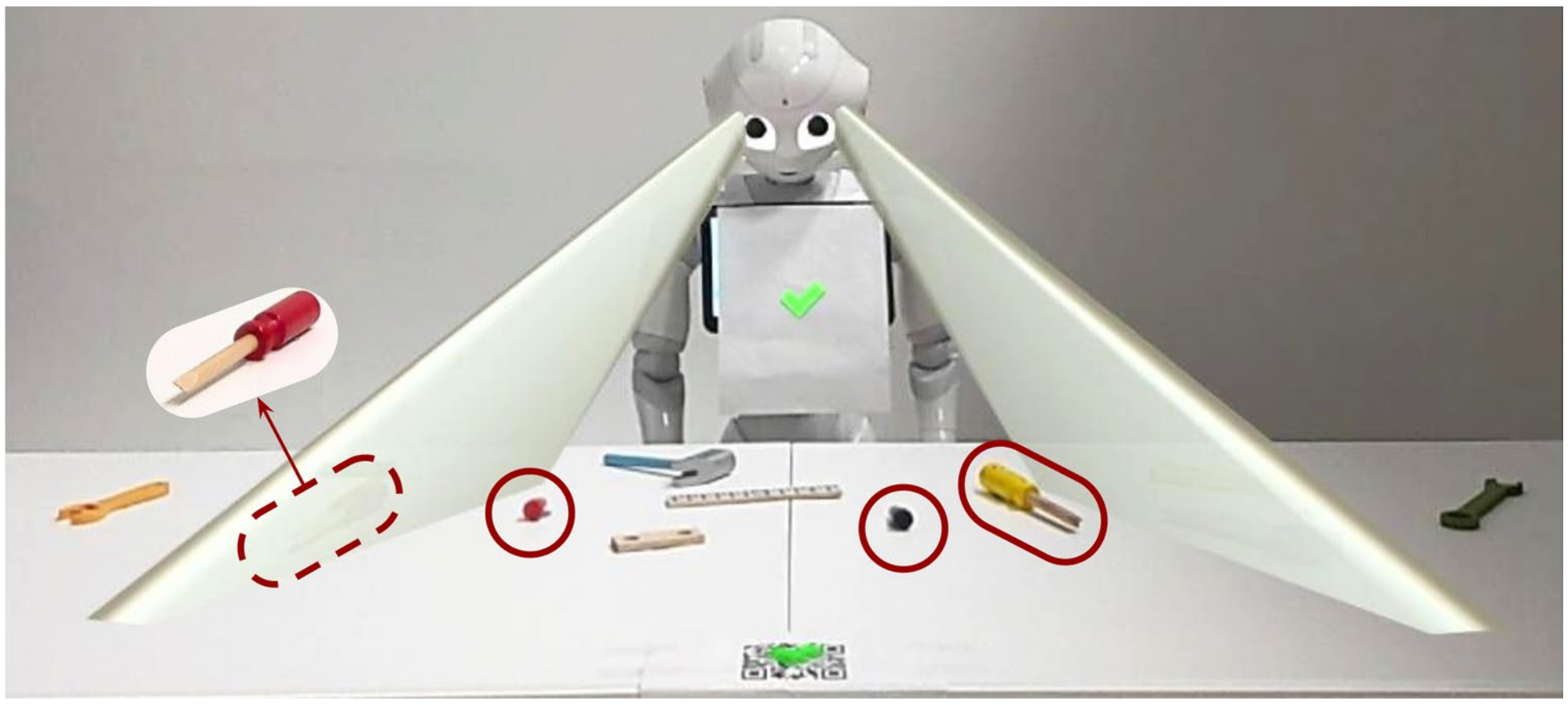

**Fig. 6** The four clustered objects needed for assembly (red circled), shown in Extended Blocks design. From left to right, red screwdriver is out of FoV. Red screw, blue screw, yellow screwdriver are within FoV. The yellow screwdriver is changed to outside-FoV half of the time by moving it to the right to mimic real-world cluster scenarios

They were asked to guess whether Pepper could see each object, i.e., whether it was within the robot's FoV. If they believed so, they said they wanted the robot to hand it. Otherwise, they said they wanted to take it themselves.

To avoid ordering effects, we used a balanced Latin Square for the ordering of the instructions and, thus, the ordering of the corresponding objects. To mimic the real-world placements of the four objects for ecological validity, we flipped the yellow screwdriver's visibility as explained in Fig. 6.

## 6.3 Implementation

For implementation, we developed all AR indicators in Unity. They precisely matched their physical dimensions and positions of the robot and the task-related objects. To register them onto the physical robot, we used the Vuforia Engine [45]'s tracking capability by attaching a QR code on the robot's chest screen. To achieve visual coherence, we attached an invisible phantom model of the robot's head to disable rendering the part of the AR indicators occluded by the robot's physical head. To register the other ends of Extended Blocks and Blocks at Task to the table, we placed another QR code in the middle of the table.

As shown in Fig 7, we also implemented a menu for HoloLens 2 using Unity. It has different buttons to switch between different indicators. A participant can choose an indicator by selecting with a finger, and the indicators will be displayed on the robot or in the task environment after scanning the QR codes.

## 6.4 Procedure

Upon arrival, each participant completed an informed consent form. Once agreed to participate, they completed a demographic survey and watched three videos to learn how to wear HoloLens 2 [43], how to choose different designs by pointing through buttons and how to scan the QR codes [2], and how to read the instructions [1]. Experimenters then briefly reintroduced the task and asked clarification questions.

Next, participants scanned the QR codes on the robot's chest and the table to register the designs. While facing away from the robot, they sat on a wheeled chair and read a page to understand the assembly goal and then read the following assembly instruction page. Once ready, they pressed the numbered button on the AR menu (Figure 7) corresponding to the condition assigned by the experimenter and turned to face the robot to start the task. The full condition name was abbreviated to avoid influencing participants' decisions on what the robot can see. To ensure participants see the designs fully, participants sat approximately 3.3 meters away from the robot and were asked to move the chair along the 1.82-meter-wide table [38] to check the design from multiple perspectives. If they believed the robot could see the object, they said they wanted the robot to hand it. Otherwise, they said they wanted to take it themselves. After each condition, they filled out the confidence and workload questionnaires. They repeated this process until all four instructions were done and the airplane model was assembled. While finishing each step, experimenters did not reveal participants whether the robot could actually see

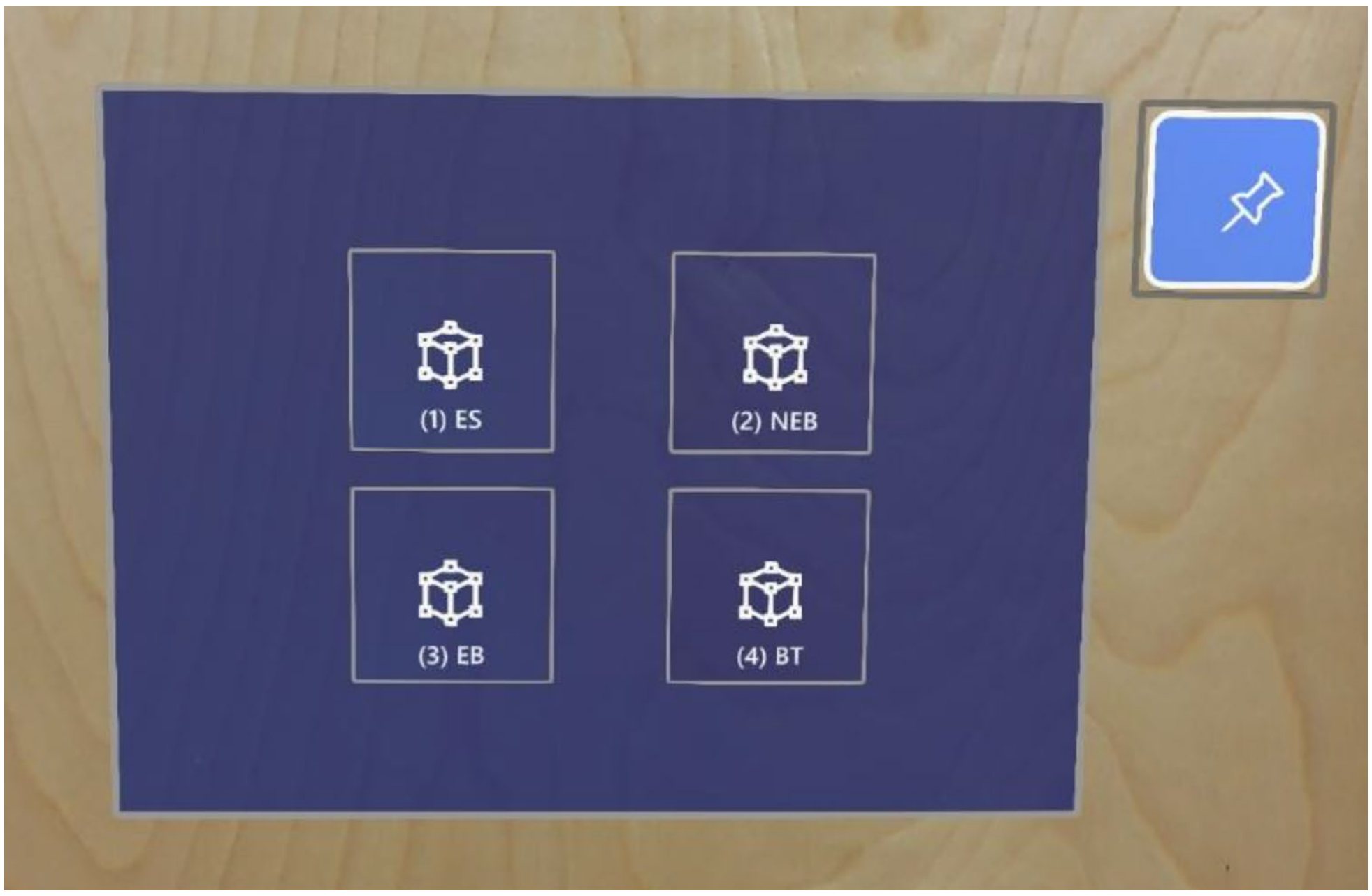

**Fig. 7** AR menu interface with four design buttons for participants to select. The four buttons are labeled only with numerical IDs and short abbreviations (ES, NEB, EB, BT). Participants were told which number to press for each trial and were not told what the abbreviations stand for to avoid influencing their decision on what the robot can see from the full condition names

the tool. This is to prevent participants from knowing the robot's FoV.

It took an average of 30:18 minutes to finish the study, and each participant was paid US$10 gift card as compensation.

## 6.5 Data Collection and Measures

**Accuracy** was calculated as the percentage of correct requests among all requests, whether they correctly guessed the object within or outside FoV. **Instruction completion time** was coded from the videos frame by frame from when the participants turned around to face the robot to when they said either they wanted the robot to get the tool or wanted to get it themselves. After outlier analysis, we removed intervals of more than 30 seconds, an excessive amount of time for a guess. The distribution of further details on the outlier analysis can be found in Appendix A.1. For the NASA Task Load Index [20, 39] measuring **cognitive effort**, we used both the load survey and its weighting component to calculate a weighted score. In the seven-point Likert scale to measure **confidence**, participants were asked how confident they believed the robot could see the object: "I was confident that the robot can see the tool or the object needed". We reversed the scores if they wanted to get the object themselves. Additionally, we asked a **free-response question** to seek qualitative feedback for them to explain their responses.

For completion time, two coders coded the videos frame by frame for the start and end of following an instruction. They jointly coded a random 10% of the videos and the rest 90% were coded solely by the other coder. Because the videos were shot at 30 frames per second, the inter-rater agreement depends on the allowable frame difference chosen. Shown in Figure 8, we achieve a $\kappa$ value over 0.8 (almost perfect agreement [32]) when the frame difference is 9 (0.3 seconds).

## 6.6 Data Analysis

Our data analysis used a Bayesian analysis framework [55], which allows us to quantify evidence for and against competing hypotheses, including the null hypothesis ($\mathcal{H}_0$). Unlike the Frequentist approach, which cannot provide evidence in favor of $\mathcal{H}_0$, the Bayesian method uses the Bayes Factor (BF) to compare the likelihood of data under two competing hypotheses: $\mathcal{H}_1$ ($\bar{x_1} \neq \bar{x_2}$, presence of an effect) and $\mathcal{H}_0$ ($\bar{x_1} = \bar{x_2}$, absence of an effect). For instance, $BF_{10} = 5$ means that the data is five times more likely to occur under $\mathcal{H}_1$ than $\mathcal{H}_0$, thus supporting $\mathcal{H}_1$.

We also used a *credible* interval (CI) instead of Frequentist's confidence interval, a random interval that contains the estimated parameter of $\gamma$% of the time. A credible interval provides a direct probability statement, i.e., $\alpha$% probability that the parameter would fall in the interval.

To interpret the results of our Bayes Factor analyses, we used the widely accepted discrete classification scheme proposed by [33]. For evidence favoring $\mathcal{H}_1$, a Bayes factor $BF_{10}$ is deemed anecdotal (inconclusive) when $BF_{10} \in (1,3]$, moderate when $BF_{10} \in (3,10]$, strong when $BF_{10} \in (10,30]$, very strong when $BF_{10} \in (30,100]$, and extreme when $BF_{10} \in (100,\infty)$. Anecdotal evidence is considered inconclusive while others are conclusive.

In the opposite, for evidence favoring $\mathcal{H}_0$, i.e., against $\mathcal{H}_1$, the intervals are inverted: Anecdotal (inconclusive) when $BF_{01} \in (1,3]$, moderate when $BF_{01} \in (3,10]$, strong when $BF_{01} \in (10,30]$, very strong when $BF_{01} \in (30,100]$, and extreme when $BF_{01} \in (100,\infty)$.

For frequency data, we ran Bayesian multinomial and post hoc binomial tests. We also ran Bayesian repeated measures ANOVA tests to analyze the repeatedly measured conditions, i.e., designs {0,1,2}, {0,1,2,3} and {0,1,2,4} because participants only experienced one allocentric design (design 3 or 4) after all egocentric designs. When $BF_{10}$ or $BF_{01} \in$ [1, 3] or (i.e., inconclusive), we ran post hoc t-tests for pairwise comparisons. For designs 3 and 4, we ran an independent sample t-test.

## 6.7 Participants

41 participants were recruited from the authors' institution through flyers. In a free-form response question, 27 (66%) identified as male, 14 (34%) identified as female, and none reported other gender identities. Age ranges from 18 to 30 ($M$=21, SD=2.9). For racial data, they were about half

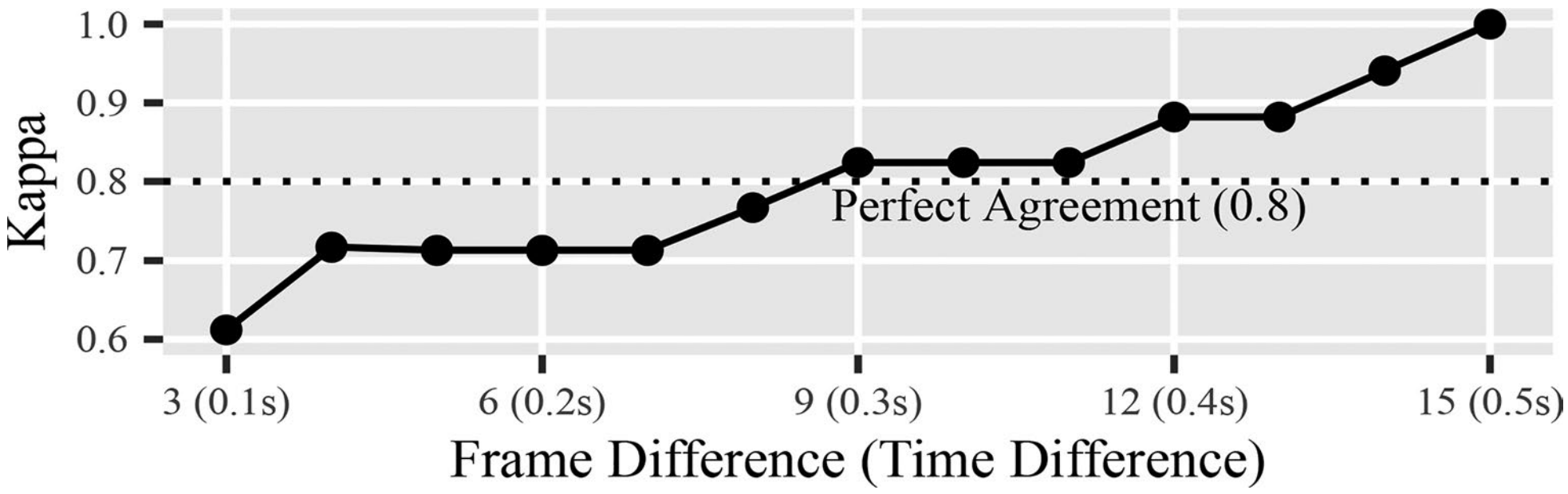

**Fig. 8** Cohen's $\kappa$ values for video coding show above perfect inter-coder agreement [32] when frame difference is 9 (0.3s)

**Table 2** Means and standard deviations (SD) for all measures across all conditions

| Measure | Baseline | Eye Sockets | Near-Eye Blocks | Extended Blocks | Blocks at Task |
|---|---|---|---|---|---|
| Accuracy | 66% | 85% | 71% | 81% | 95% |
| Completion Time (s) | $9.615 \pm 6.565$ | $10.978 \pm 7.997$ | $9.545 \pm 6.408$ | $6.550 \pm 3.238$ | $11.418 \pm 6.235$ |
| Confidence | $5.732 \pm 1.073$ | $5.610 \pm 1.202$ | $5.317 \pm 1.572$ | $6.190 \pm 1.569$ | $5.850 \pm 1.137$ |
| Cognitive Effort | $24.244 \pm 17.095$ | $25.439 \pm 17.496$ | $22.675 \pm 17.058$ | $22.778 \pm 16.503$ | $20.400 \pm 17.654$ |

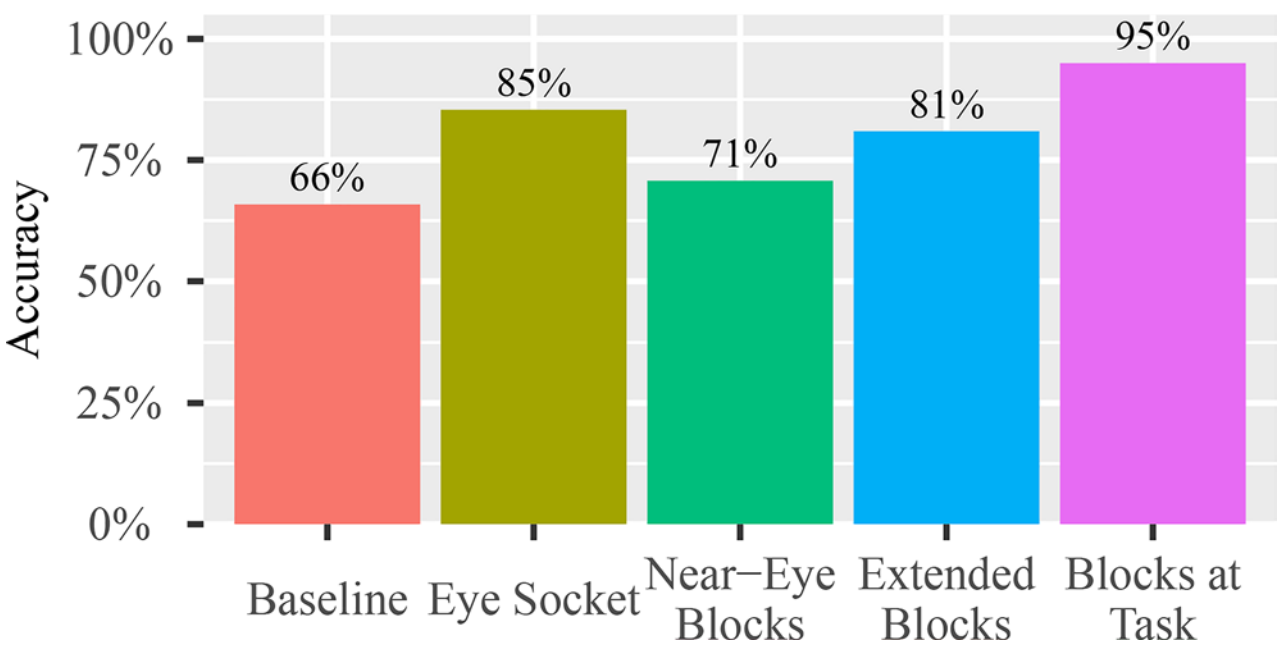

**Fig. 9** Accuracy percentages across different conditions, showing the proportion of correctly made requests to the robot to hand over objects within its FoV relative to total requests made. The Blocks at task condition is the most accurate

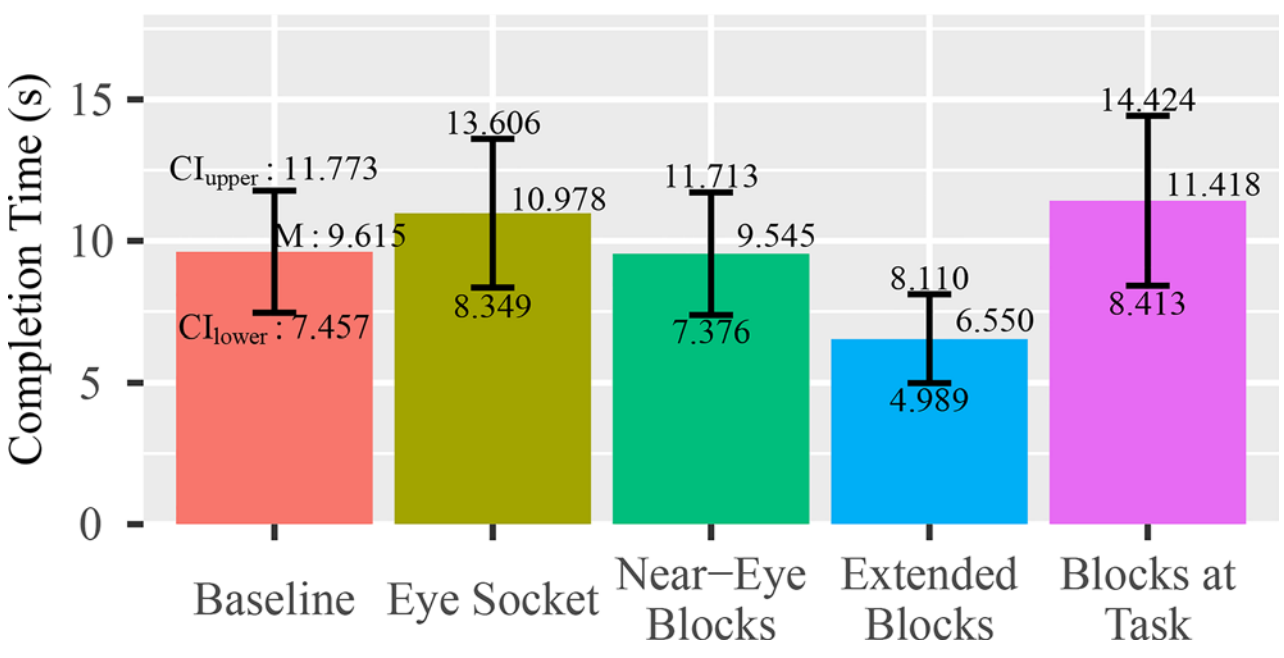

**Fig. 10** Mean completion time across different conditions. Error bars show 95% credible interval (CI): baseline [7.457, 11.773], Eye Socket [8.349, 13.606], Near-Eye Blocks [7.376, 11.713], Extended Blocks [4.989, 8.11], Blocks at task [8.413, 14.424]. Strong evidence ($BF_{10}$ = 14.242) favors a difference between Extended Blocks and Blocks at task. Moderate evidence ($BF_{01} \in [4.161, 7.513]$) favors no differences among other pairwise comparisons

Asian (19, 46.3%) and one-third White (13, 31.7%), while five (12.2%) were Latino/Hispanic identities, two (4.9%) were Black, and two (4.9%) reported multi-racial. Experience with robots and AR was measured on 7-point Likert items (‘I have experience using robots’; ‘I have experience using augmented reality’). For robots, 21 participants (51.2%) agreed (ratings 5–7), six (14.6%) were neutral (rating 4), and 14 (34.1%) disagreed (ratings 1–3). For AR, 23 (56.1%) agreed, five (12.2%) were neutral, and 13 (31.7%) disagreed.

## 7 Results

We ran all Bayesian tests in an open-source statistics program JASP 0.19.0 [22]. Table 2 summarizes the means and standard deviations for accuracy, completion time, confidence, and cognitive effort across all conditions.

In brief, all indicators improved accuracy over the Baseline, with Blocks at Task achieving the highest accuracy (95%) and Eye Sockets also performing well (85%). Extended Blocks yielded the shortest completion times on average, whereas Blocks at Task took longer despite its high accuracy. Confidence ratings were generally high (around 5.3–6.2 on a 7-point scale) and showed no differences between conditions. NASA-TLX workload scores were low across all designs (around 20–25 on a 0–100 scale) and did not differ between conditions. In the following subsections, we report the detailed Bayesian analyses.

### 7.1 Accuracy

As shown in Figure 9, accuracy ranges from 66% to 95%. We first conducted a Bayesian multinomial test [15] on the frequency data with equal proportions, revealing extreme evidence ($BF_{10} = 8.582 \times 10^6$) favoring an effect. We thus conducted post-hoc Bayesian binomial tests [41] with one-sided alternative hypotheses ($\mathcal{H}_1$) that the proportion is larger than 50% for all conditions.

Results showed an inconclusive anecdotal evidence ($BF_{10} = 2.907$) favoring $\mathcal{H}_1$ in Baseline. This suggests that there probably is an effect, but if there was, it would be that the data is 2.907 times likely under $\mathcal{H}_1$, but more data would be needed to fully confirm such an effect. Otherwise, results revealed conclusive evidence favoring $\mathcal{H}_1$ for all the frequency data in all other conditions: Extreme evidence for Eye Socket ($BF_{10} = 23288.748$), strong evidence for Near-Eye Blocks ($BF_{10} = 13.205$), very strong evidence for Extended Blocks ($BF_{10} = 31.785$), and extreme evidence for Blocks at Task ($BF_{10} = 4993.167$).

### 7.2 Completion Time

The mean completion time (Figure 10) ranges from 6.55 to 11.418 seconds: Baseline ($M$=9.615, SD=6.565, 95% CI: 7.457, 11.773), Eye Socket ($M$=10.978, SD=7.997, 95% CI: 8.349, 13.606), Near-Eye Blocks ($M$=9.545, SD=6.408, 95% CI: 7.376, 11.713), Extended Blocks

($M$=6.550, SD=3.238, 95% CI: 4.989, 8.110), and Blocks at Task ($M$=11.418, SD=6.235, 95% CI: 8.413, 14.424).

As we planned to use Bayesian repeated measures ANOVA (RM-ANOVA) [50], we assessed the normality of the data by inspecting the Q-Q (Quantile-Quantile) plots of all conditions, a well-accepted practice among Bayesianists [55]. As we found violations of normality and linearity, we log-transformed the data and successfully addressed them.

A Bayesian RM-ANOVA on designs {0,1,2} revealed moderate evidence ($BF_{01}$=4.161) favoring the null hypothesis $\mathcal{H}_0$, which means there is no difference among Baseline, Eye Socket, and Near-Eye Blocks in completion time. Comparing Extended Blocks with the first three designs, a Bayesian RM-ANOVA on designs {0,1,2,3} showed moderate evidence ($BF_{01}$=7.513) against an effect of completion time among Baseline, Eye Socket, Near-Eye Blocks and Extended Blocks. Similarly, when comparing Blocks at Task with the first three designs, a Bayesian RM-ANOVA on design {0,1,2,4} revealed moderate evidence ($BF_{01}$=5.375) against an effect of completion time among Baseline, Eye Socket, Near-Eye Blocks, and Blocks at Task.

Finally, for comparison between Extended Blocks ($M$=6.550; 95% CI: 4.989, 8.110) and Blocks at Task ($M$=11.418; 95%CI: 8.413, 14.424), a Bayesian independent samples t-test revealed strong evidence favoring a difference (4.868; $BF_{10}$=14.242).

### 7.3 Confidence

As shown in Figure 11, mean confidence ratings range from 5.32 to 6.19 out of 7: Baseline ($M$=5.732, SD=1.073, 95% CI: 5.393, 6.070), Eye Socket ($M$=5.610, SD=1.202, 95% CI: 5.230, 5.989), Near-Eye Blocks ($M$=5.317, SD=1.572, 95% CI: 4.821, 5.813), Extended Blocks ($M$=6.190, SD=1.569, 95% CI: 5.476, 6.905), and Blocks at Task ($M$=5.850, SD=1.137, 95% CI: 5.318, 6.382).

A Bayesian RM-ANOVA on designs {0,1,2} revealed moderate evidence ($BF_{01}$=3.915) against any difference in confidence among Baseline, Eye Socket, and Near-Eye Blocks. When comparing Extended Blocks with the first three designs, a Bayesian RM-ANOVA on designs {0,1,2,3} showed moderate evidence ($BF_{01}$=3.418) against an effect of confidence among Baseline, Eye Socket, Near-Eye Blocks, and Extended Blocks. Similarly, when comparing Blocks at Task with the first three designs, a Bayesian RM-ANOVA on designs {0,1,2,4} showed moderate evidence ($BF_{01}$=3.582) against an effect of confidence among Baseline, Eye Socket, Near-Eye Blocks, and Blocks at Task.

Finally, a Bayesian independent samples t-test on confidence between Extended Blocks and Blocks at Task revealed anecdotal evidence ($BF_{01}$=2.545) favoring no difference, suggesting that there probably is no such effect, but if there was, it would be that participants would be 2.545 more likely to be equally confident in Extended Blocks and Blocks at Task. More data would be needed to fully rule out such an effect.

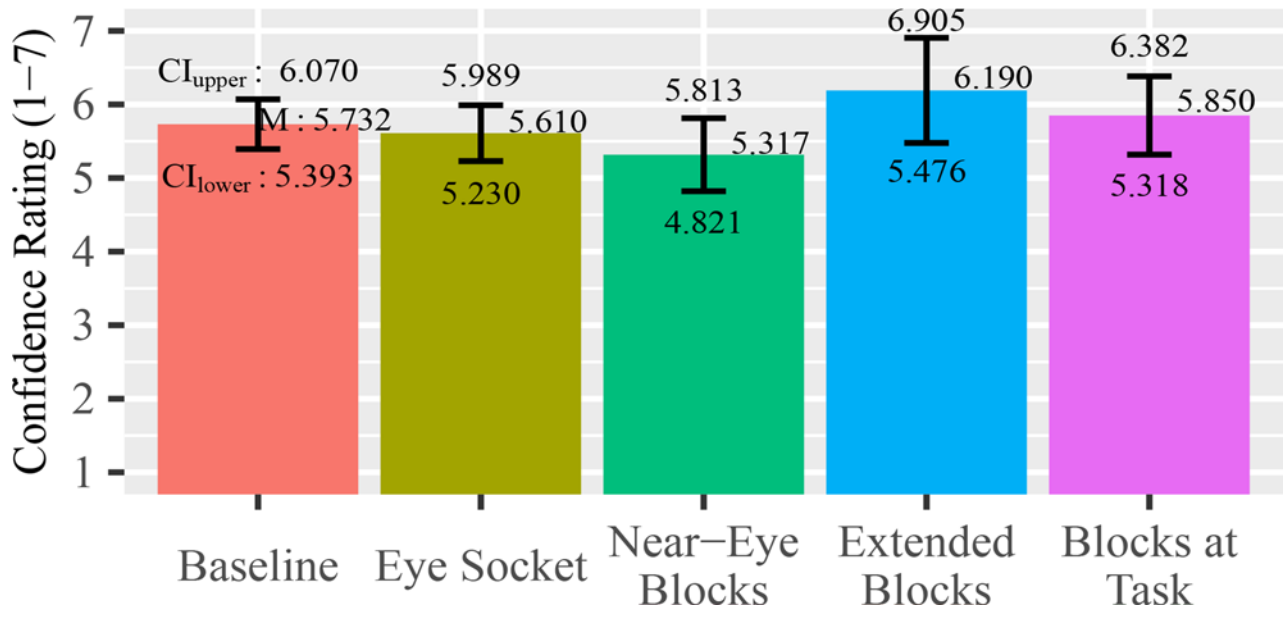

**Fig. 11** Mean confidence rating across different conditions. Error bars show 95% CI: baseline [5.393, 6.070], Eye Socket [5.230, 5.989], Near-Eye Blocks [4.821, 5.813], Extended Blocks [5.476, 6.905], Blocks at task [5.318, 6.382]. Results favor no differences among all pairwise comparisons ($BF_{01} \in$ [3.418, 3.915]) except for Extended Blocks and Blocks at task ($BF_{01}$=2.545)

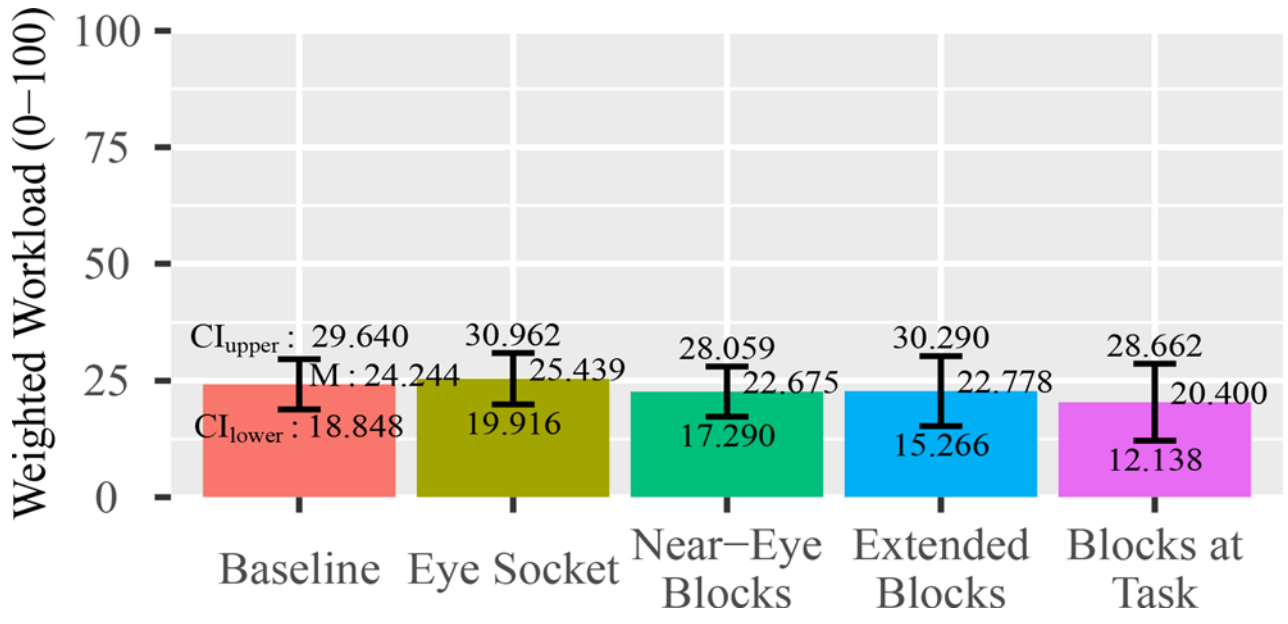

**Fig. 12** Mean weighted workload. Error bars show 95% CI: baseline [18.848, 29.640], Eye Socket [19.916, 30.962], Near-Eye Blocks [17.290, 28.060], Extended Blocks [15.266, 30.290], Blocks at task [12.138, 28.662]. Results favor no differences among all pairwise comparisons ($BF_{01} \in$ [3.019,5.368]), except for Extended Blocks and Eye Socket ($BF_{01}$=1.406)

### 7.4 Cognitive Effort

Shown in Figure 12, mean workloads are low and range from 20.4 to 25.4 in 100: Baseline ($M$=24.244, SD=17.095, 95% CI: 18.848, 29.640), Eye Socket ($M$=25.439, SD=17.496, 95% CI: 19.916, 30.962), Near-Eye Blocks ($M$=22.675, SD=17.058, 95% CI: 17.290, 28.059), Extended Blocks ($M$=22.778, SD=16.503, 95% CI: 15.266, 30.290), and Blocks at Task ($M$=20.400, SD=17.654, 95% CI: 12.138, 28.662).

A Bayesian RM-ANOVA on designs {0,1,2} revealed moderate evidence ($BF_{01}$=5.368) against any difference among Baseline, Eye Socket, and Near-Eye Blocks. Comparing Extended Blocks with the first three designs, a Bayesian RM-ANOVA on designs {0,1,2,3} showed anecdotal evidence ($BF_{01}$=2.91) against a difference among Baseline,

Eye Socket, Near-Eye Blocks, and Extended Blocks. We thus ran post-hoc t-tests that revealed moderate evidence ($BF_{01}=4.186$) against a difference between Extended Blocks and Baseline, anecdotal evidence ($BF_{01}=1.406$) between Extended Blocks and Eye Socket, and moderate evidence ($BF_{01}=4.043$) between Extended Blocks and Near-Eye Blocks. Comparing Blocks at Task with the first three designs, a Bayesian RM-ANOVA on designs {0,1,2,4} showed moderate evidence ($BF_{01}=4.865$) against a difference among Baseline, Eye Socket, Near-Eye Blocks, and Blocks at Task.

Finally, for comparison between Extended Blocks and Blocks at Task, a Bayesian independent samples t-test revealed moderate evidence ($BF_{01}=3.019$) against a difference in workload between Extended Blocks and Blocks at Task.

# 8 Discussion

## 8.1 Hypothesis One: Accuracy

Our first hypothesis was that as the indicators got closer to the task space, participants would develop a more accurate mental model of the robot's FoV. Shown in Figure 9, H1 was mostly supported except for Eye Socket, which was also relatively accurate (85%).

Without any FoV indicators, Baseline accuracy was only 66%, indicating that about one in three participants misunderstood the range of the robot's FoV and had a wrong mental model of the robot's vision capabilities. The simple Near-Eye Blocks also did not help: about 30% participants made the wrong guesses. Although functionally blocking the robot's FoV, it was not transferred to the task space, i.e., where it lies within or out of FoV on the table.

Surprisingly, the Eye Socket design is more accurate than Near-Eye Blocks and Extended Blocks. This may be because the deepened eye socket is more natural and human-like, serving a familiar reference point to participants' own eyes, allowing them to imagine the robot's limited vision range from their own. As the more accurate Blocks at Task design must be AR, we propose **Design Guideline 1: Without other AR indicators, robot designers should design deeper eye sockets to match each camera's FoV**.

The Extended Blocks design had a lower accuracy rate (81%) than Blocks at Task (95%). After analyzing the free-form responses from the four participants who were wrong, we found that triangle-shaped panels were perceived as two 3D cones for its peripheral vision projected from eye sides (P25: *"I could see his vision more clearly with the simulated cones"*), and thought the robot could only see the objects within the cones (P13: *"...the robot could see it because the flare illuminated the red screwdriver."*). For the out-of-view tools occluded by the panels but appeared inside the cones, they thought that these objects were within the robot's FoV. For the tools within FoV but not in the cones, they believed they were out of the FoV (P25: *"the screwdriver was not within the cone, so I assume he could not see it."*).

The triangle-panel-to-cone misconception reveals a problem with optical see-through AR devices like HoloLens 2, where the virtual content is light reflected onto the optical lenses, appearing semi-transparent, and, thus, cannot fully occlude physical objects (i.e., light cannot block light). That is, participants can still see the objects through the AR panels (P21: *"the robot had the desired tool highlighted"*), and therefore incorrectly thought those tools were covered by the simulated cones, leading them to the wrong decision.

One solution is to use video see-through devices like Apple Vision Pro, instead of optical see-through devices, so those out-of-FoV objects can be fully occluded. Another solution is to use rectangular blocks rather than triangle blocks that people treat as cones, or the Blocks at Task design solely in the task space. Future work should examine these solutions.

To conclude, our accuracy results showed that the indicator at the task space helped understand the robot's vision capabilities the most. Thus, we propose **Design Guideline 2: If AR situated visualization can be leveraged, robot designers should add FoV indicators at the task space for nearly perfect accuracy.**

## 8.2 Hypothesis Two: Task Efficiency

Our second hypothesis was that indicators closer to the task space would enhance task efficiency, measured by completion time. H2 is almost unsupported: Results favor no difference among all pairs, except for strong evidence supporting a difference between Extended Blocks and Blocks at Task.

Compared with Extended Blocks, Blocks at Task is a task-centric allocentric design, which disconnects, or lacks transition, from the eyes. We observed some participants spend time connecting this design back to the robot's FoV. They were thinking for a while about the use of this design when they first saw them. P18 explained the connection process, *"I didn't think those walls would be there. This added ... uncertainty. I ... modeled my arms as the walls. I couldn't see my screwdriver, so I assumed the robot couldn't see it's screwdriver either."* This may explain why participants spent five seconds fewer in Extended Blocks that connect back to eyes. Indeed, other AR works within HRI had similar findings, e.g., robots referring to objects by AR circles delayed completion time due to the connection process despite being more accurate [8].

Although participants spent more time on Blocks at Task, the accuracy was the highest. Thus, we still retain Design Guideline 2. However, with the efficiency benefit, we propose **Design Guideline 3: Robot designers should connect AR FoV indicators at the task space to the eyes for efficiency.**

## 8.3 Hypothesis Three: Confidence

Our third hypothesis was that indicators closer to the task space would enhance confidence in gauging the robot's FoV. H3 is almost unsupported: Results favor no difference among all pairs except for anecdotal evidence against a difference between Extended Blocks and Blocks at Task. This indicates that proximity to the task environment did not affect confidence, reinforcing design guidelines 1 and 2.

We conducted an additional analysis of confidence levels among participants who made incorrect decisions. Under Extended Blocks, those who were wrong were still highly confident, scoring 6.5 out of 7. This suggests that Extended Blocks led to overconfidence in incorrect assumptions. In contrast, Baseline, Eye Socket, and Near-Eye Blocks had lower confidence in their wrong guesses, 5.57, 5.5, and 5.25, respectively (As only one participant was wrong in Block at Tasks, we omitted its confidence value). These numbers roughly match the overall confidence shown in Fig. 11. Thus, we propose **Design Guideline 4: If Extended Blocks is used alone, robot designers should be aware that wrong guessers might be overconfident.**

## 8.4 Hypothesis Four: Workload

Our last hypothesis was that designs closer to the environment would reduce cognitive effort. H4 is almost unsupported: Results favor no difference among all pairs except for anecdotal evidence against a difference between Extended Blocks and Eye Socket.

Results showed low workloads in all conditions, capped at 25.4/100. This also includes Blocks at Task: Although participants spent more time guessing, the workload has not increased. A likely reason for the low workload scores across all indicators is that guessing the FoV itself is not a demanding task, although different FoV indicators made differences in the previous three aspects just discussed like completion time. Thus, we propose **Design Guideline 5: Robot designers should rest assured that although the highly accurate FoV indicator at the task space has lower task efficiency, the workload has remained low.**

## 8.5 General Discussion

Generally, our findings showed three designs helped address people's misunderstanding about a robot's FoV. For allocentric designs at the task space, Blocks at Tasks is the most accurate but at a completion time cost. Extended Blocks is promising but the triangle-panel-to-cone misconception needs to be solved, after which it will combine both accuracy and efficiency benefits. For egocentric designs at the eyes and head space, Near-Eye Blocks did as bad as Baseline, while simple Eye Socket deepening, providing cues about its FoV possibly by physical alteration, improved accuracy.

Finally, based on the results, we propose an **application-specific Design Guideline 6: For mission-critical collaborative tasks that require accuracy, the allocentric design like Blocks at Task should be used**.

## 8.6 Limitations and Future Work

Besides the limitation of task design on workload, we focused on addressing the misunderstanding of robots' *horizontal* FoV. Yet, robots also have different vertical FoV (e.g., Pepper's 44.6° [4], Fetch's 45° [47, 61]) than human's (160°). Although we tend to have a 2D workspace like a table at a fixed height, this discrepancy is similarly problematic as human-robot collaboration happens in a narrow workspace where, e.g., robots need to work near a multi-shelf organizer. In those cases, people will expect robots to see objects on multiple shelves while the robot can only see one or two shelves. Future research should investigate this as they are common in industrial scenarios like warehouses or factories.

Secondly, thanks to AR situated visualization, there is a growing interest in leveraging AR for HRI [19, 23, 34, 54, 56]. However, AR devices may not always be available. While the first two designs can be incorporated by physical alteration or addition, we have started exploring familiar body language as FoV indicators [59]. As shown in Figure 13, they are an egocentric *Near-Eye Hands* design, raising hands directly to the sides of its eyes to reveal FoV, and a transition-space allocentric *Extended Arms* design that extends both arms forward, similar to the AR Extended Blocks. Together with our Eye Socket and Near-Eye Blocks designs that also do not require AR, one can evaluate these four designs for non-AR scenarios to provide more insights. For the measures, a design preference question with an optional explanation can be asked as well as how the robot itself would be perceived is also of interest. Besides body language, a fifth design possibility is to leverage projector-based AR. Rather than head-mounted AR displays or physical alteration, this design uses an overhead projector to

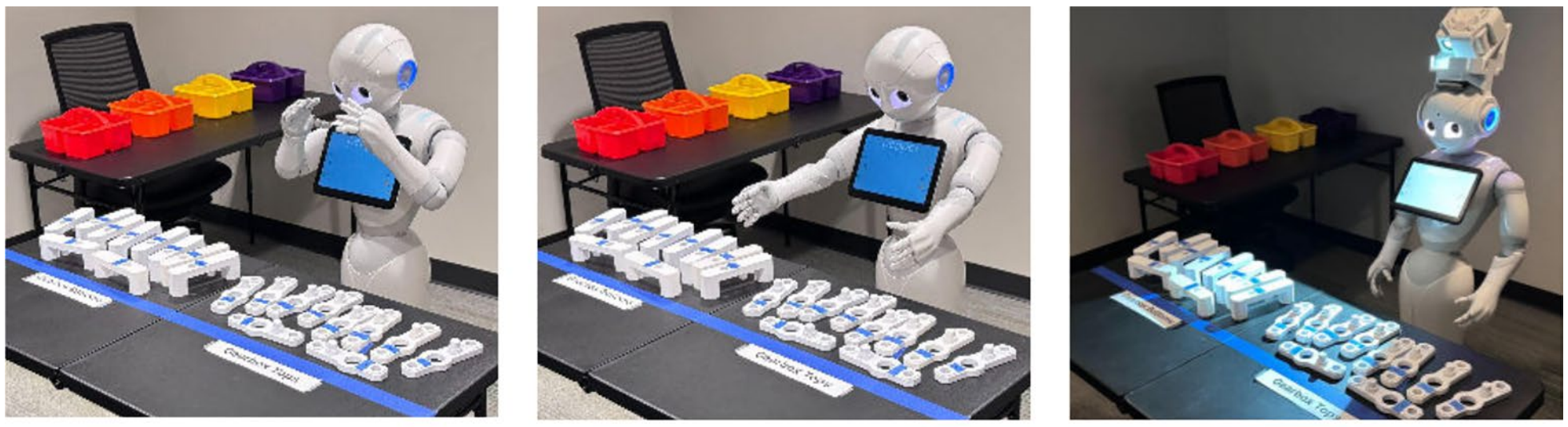

(1) Near-Eye Hands (2) Extended Arms (3) Overhead Projector

**Fig. 13** Familiar body-language indicators of a robot's FoV that we are currently exploring

project lines onto the robot's operating environment to indicate the robot's FoV. This projected AR technology frees interactants from wearing head-mounted displays or holding phones or tablets, thus making it ergonomic and scalable to a crowd, beneficial in group settings. We believe these designs will achieve higher positive perceptions as they are more familiar designs and activate human-human interaction patterns.

Thirdly, while we focused on manipulation in human-robot collaboration that often happens in controlled environments like factories, a robot may navigate and look around frequently in settings like warehouse floors, shopping malls [11], and retail stores [12]. In these contexts, a robot has more opportunity to adjust its view to overcome its limited FoV during navigation tasks. People in those more unstructured and naturalistic environments also allow investigation into spontaneous reactions and behaviors during a robot's navigation tasks. Thus, there is a knowledge gap on how navigation and the search behavior in these settings would affect people's perception of a robot's real vision capabilities. Future work with tasks in these scenarios can further expand our design guidelines.

Fourthly, our study evaluates FoV indicators along the continuum of our spectrum. It might be desirable to combine the FoV indicators at the two ends of the continuum, i.e., the pure egocentric deepened eye sockets together with the pure allocentric Blocks at Task design. The combination could further strengthen people's mental models, e.g., bringing the effectiveness benefit Blocks at Task (95% accuracy) to deepened eye sockets (95%), similar to what [8] have done to combine different forms of mixed reality deictic gestures. However, adding indicators at two distant places may distract interactants' attention. Thus, exploring such combinations and their potential benefits and tradeoffs is a promising direction for future design work.

Finally, the participant sample is disproportionately well-educated young Asian/White men (12 women and 26 men). They are more likely to have experienced robotic and AR technologies than other populations, as confirmed by our data: Both were over 50%. This gender imbalance may influence the study's conclusions and suggest a need for future research should address this disproportionality, e.g., reproducing in other cultures and involving more women to match the world average 1.07 male/female ratio at birth [3]. Additionally, similar to what we discussed in the third point regarding context, robots have a wide range of applications targeting different populations, future work would investigate, e.g., K-12 students in education settings [6] or older adults in healthcare settings [27] who may experience cognitive decline like Alzheimer's Disease and Related Dementias (ADRD) to see the age's influences on the conclusions, particularly cognitive load.

## 9 Conclusion

In this work, we designed four egocentric and allocentric AR FoV indicators, from the eye to head to the task space, and conducted a human-subjects study to investigate their performance and participants' experience in a collaborative HRI task. Confirming an inaccurate mental model from Baseline accuracy, our results showed that deeper Eye Socket, Extended Blocks, and Blocks at Task all helped align human expectations with the robot's actual FoV, enabling participants to develop a more accurate mental model of robots' vision capabilities. Results showed nearly perfect accuracy for the allocentric AR indicator of Blocks at Task and high accuracy for the egocentric Eye Socket design possible for physical alteration, while confidence and workloads are more than acceptable. We also provided concrete design guidelines on how to best apply FoV indicators that improve transparency and collaboration between humans and robots. Looking forward, our work opens new avenues for further exploration in robot transparency and expandability.

# Appendix A Additional Figures

## A.1 Completion Time Before and After Outlier Analysis

In our study, completion time measured how long participants took to judge whether the robot could see an object. To mitigate the large effects of outliers on the competition time data (see Figure A1), we excluded response times exceeding 30 seconds. Figure A1 shows completion time before outlier removal and Figure A2 presents the data after outlier removal.

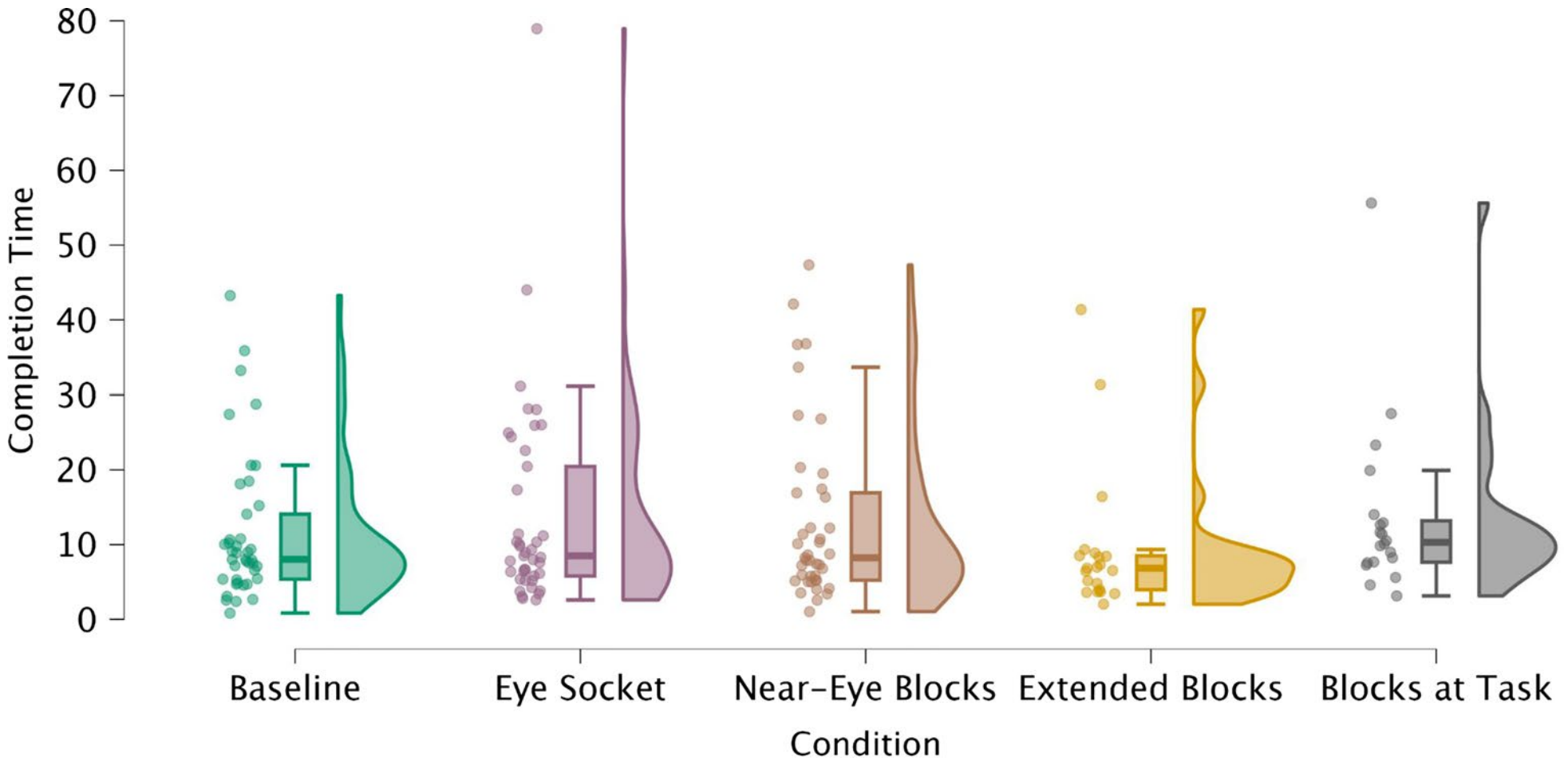

**Fig. A1** Completion time before outlier removal, showing raw distributions across all experimental conditions

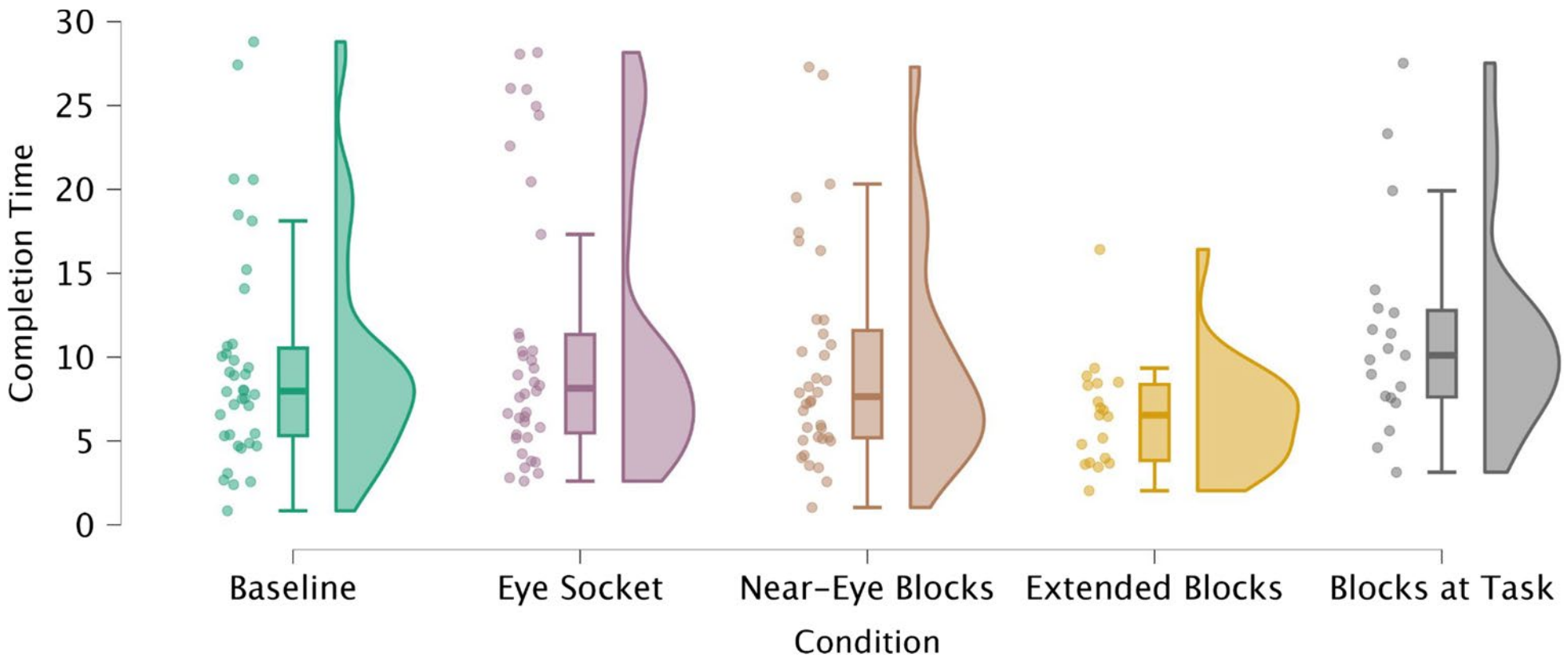

**Fig. A2** Completion time after outlier removal, filtering out responses exceeding 30 seconds

## A.2 Confidence Across Conditions

Our study followed a 1×5 design, where Baseline, Eye Socket, and Near-Eye Blocks were within-subject conditions, meaning all participants experienced these three egocentric designs. Because allocentric designs (Extended Blocks and Blocks at Task) revealed the robot's FoV, they were tested as between-subject conditions, with each participant experiencing only one of the two allocentric indicators after completing all egocentric ones.

Figures A3–A6 illustrate confidence distributions across different conditions. Figure A3 shows confidence ratings comparison among within-subject conditions (Baseline, Eye Socket, Near-Eye Blocks). Figure A4 adds the Extended Blocks condition, compare it with the three within-subject conditions. Figure A5 adds Blocks at Task condition, compare it with the three within-subject conditions. Figure A6 compares Extended Blocks and Blocks at Task.

**Fig. A3** Confidence ratings across Baseline, Eye Socket, and Near-Eye Blocks conditions

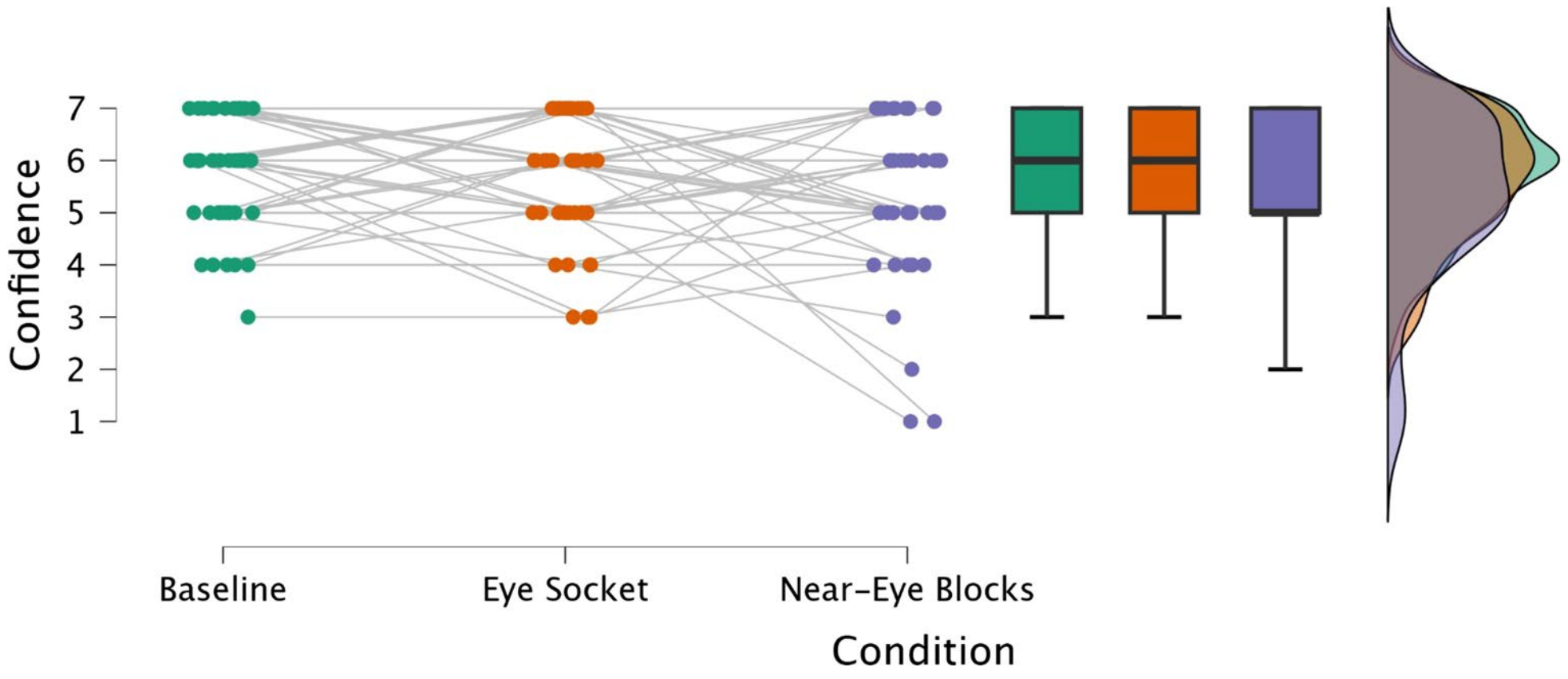

**Fig. A4** Confidence ratings across Baseline, Eye Socket, near-Eye Blocks and Extended Blocks conditions

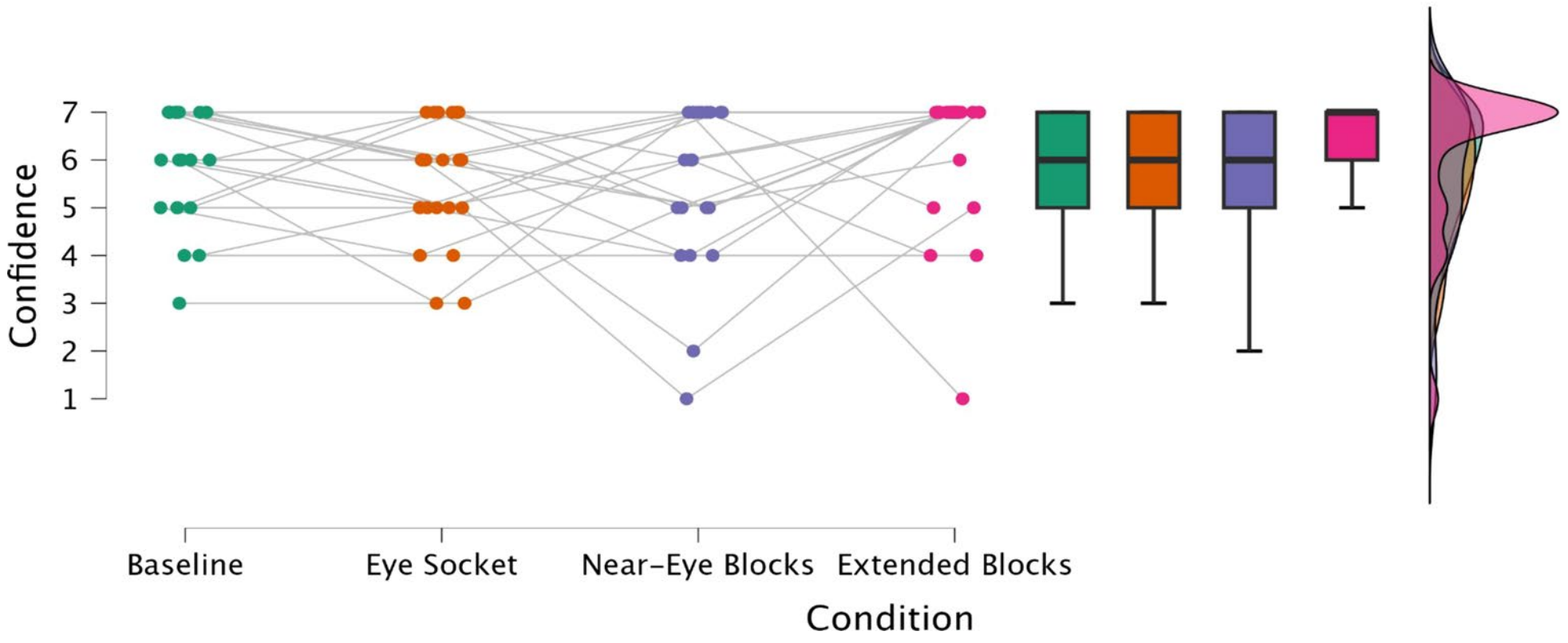

**Fig. A5** Confidence ratings across Baseline, Eye Socket, near-Eye Blocks and Blocks at task conditions

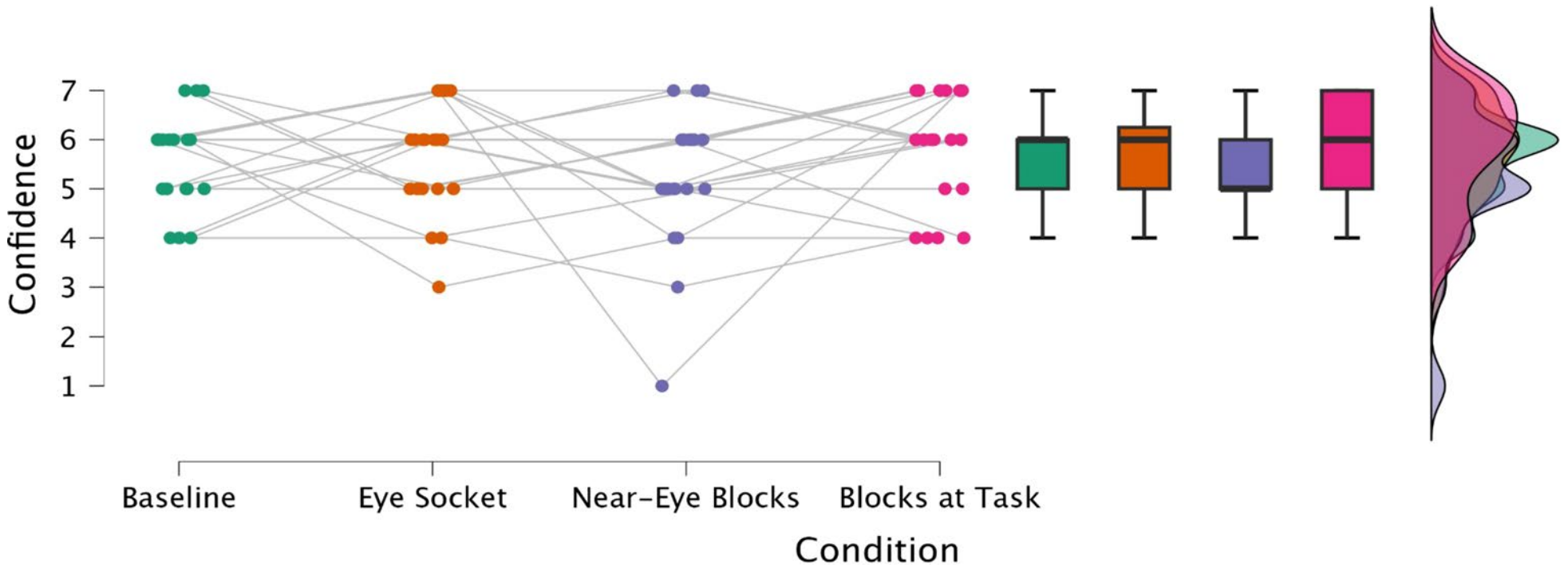

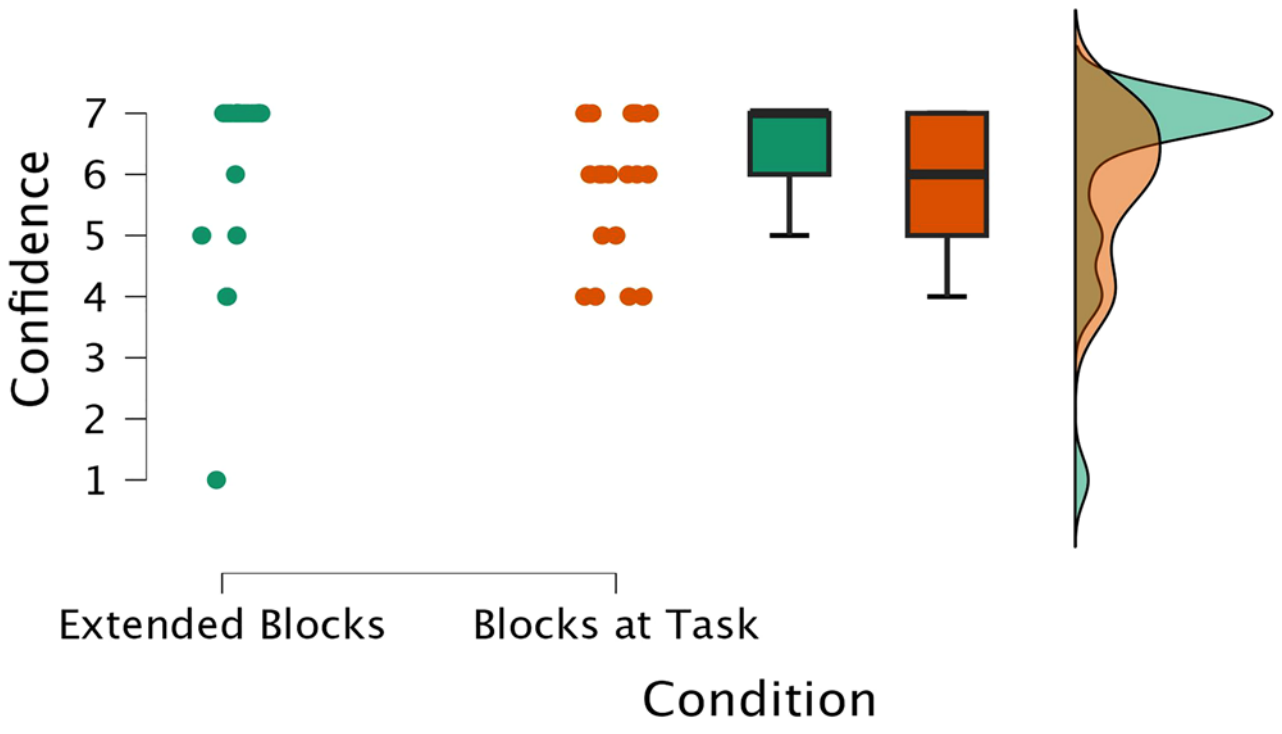

**Fig. A6** Confidence ratings for Extended Blocks vs. Blocks at task

## A.3 Cognitive Effort

Figures A7–A21 illustrate workload distributions across different conditions. Figure A7 shows workload comparison among within-subject conditions (Baseline, Eye Socket, Near-Eye Blocks). Figure A8 adds the Extended Blocks condition, compare it with the three within-subject conditions. Figure A9 adds Blocks at Task condition, compare it with the three within-subject conditions. Figure A21 compares Extended Blocks and Blocks at Task.

**Fig. A7** Workload across Baseline, Eye Socket, and Near-Eye Blocks conditions

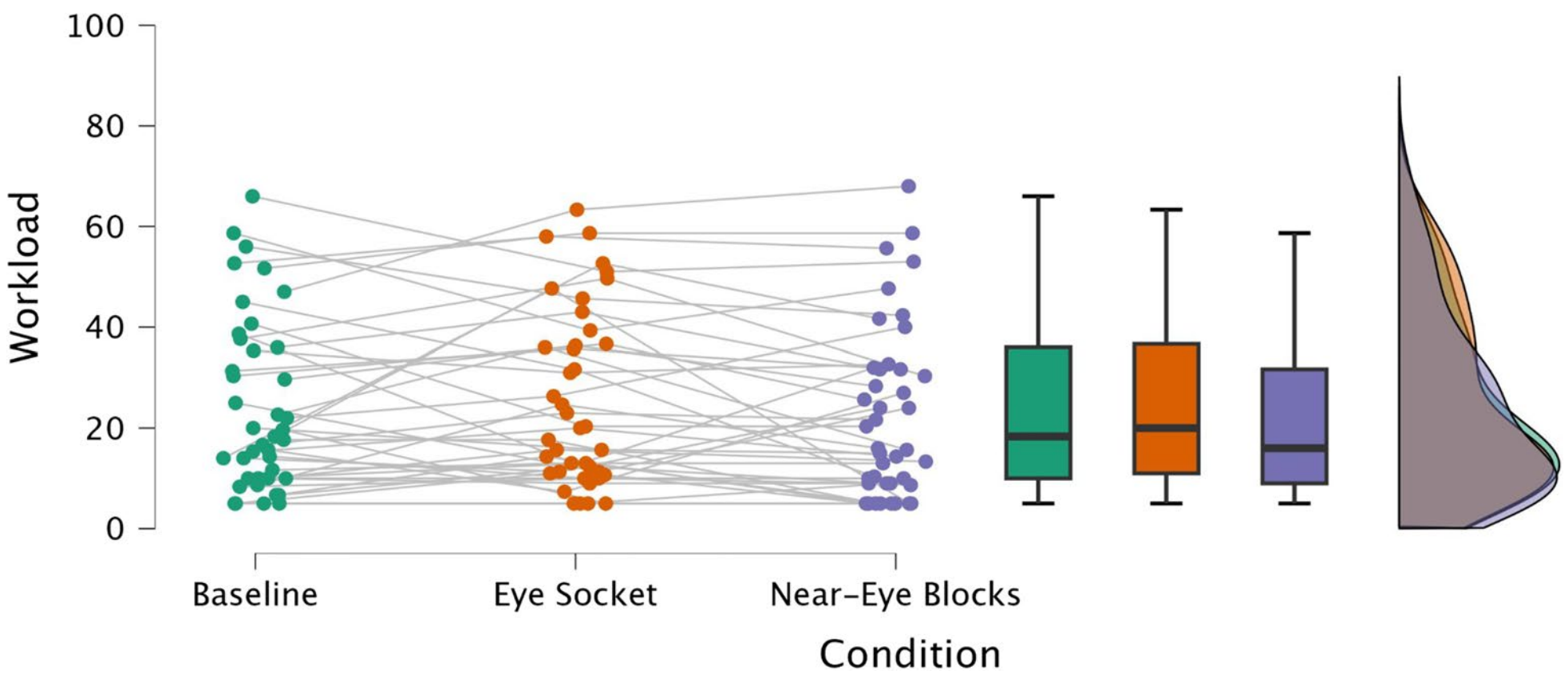

**Fig. A8** Workload across Baseline, Eye Socket, near-Eye Blocks and Extended Blocks conditions

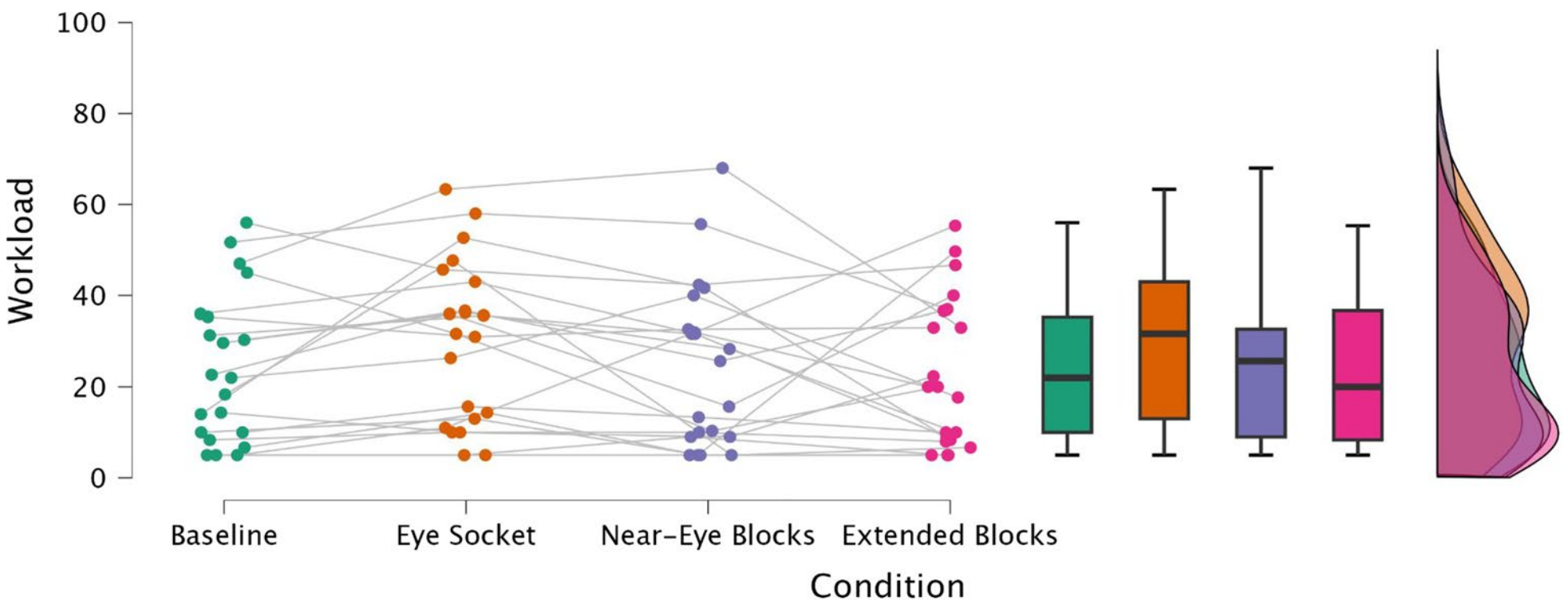

**Fig. A9** Workload across Baseline, Eye Socket, near-Eye Blocks and Blocks at task conditions

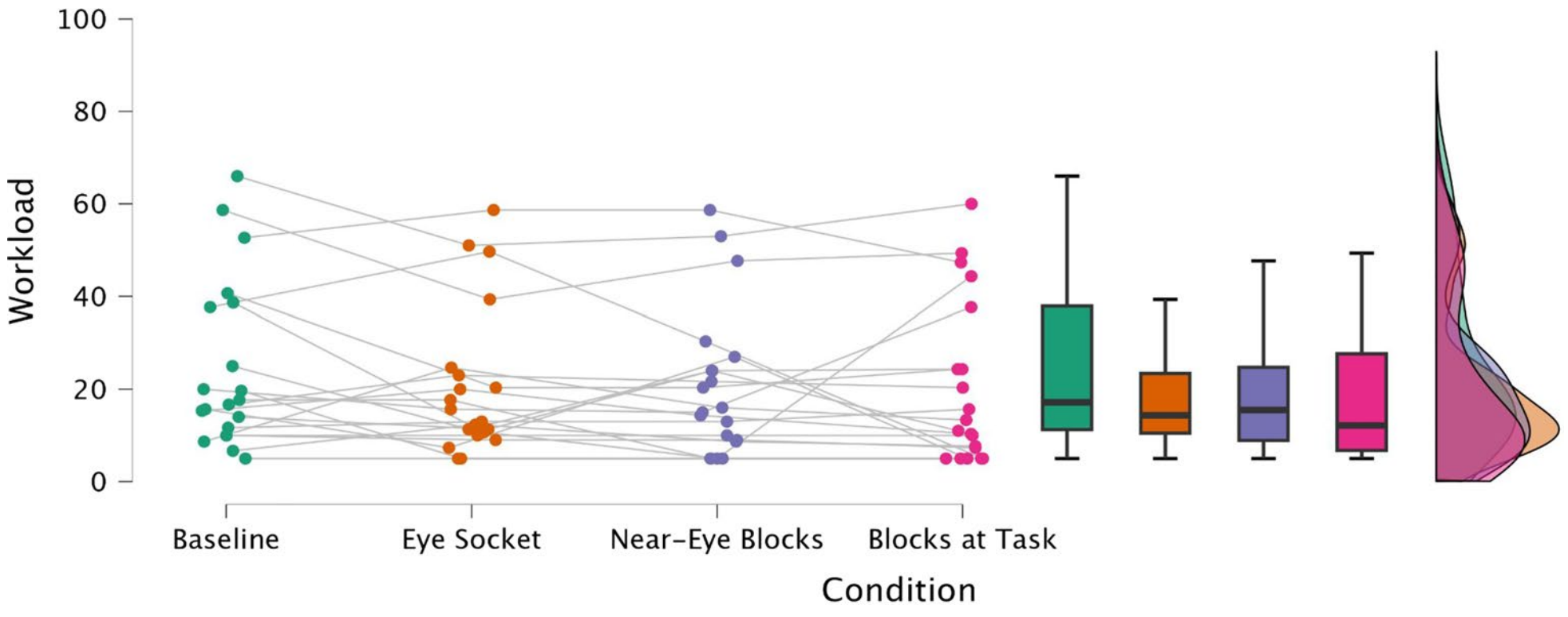

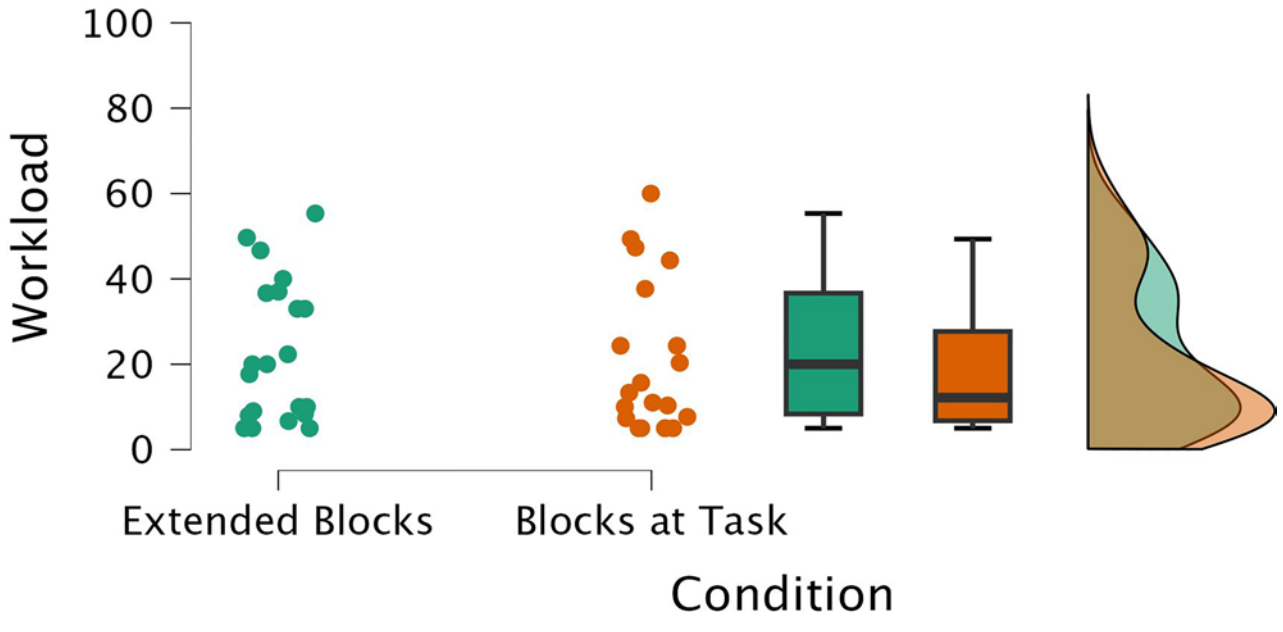

**Fig. A10** Workload for Extended Blocks vs. Blocks at task

**Acknowledgements** We thank Uthman Tijani and Xiangfei Kong for proofreading.

**Author Contributions** Conceptualization: Zhao Han, Hong Wang; Data curation: Hong Wang, Ridhima Phatak; Formal analysis: Zhao Han, Hong Wang; Software: Hong Wang, Zhao Han, James Ocampo; Methodology: Zhao Han, Hong Wang, Ridhima Phatak; Investigation: Hong Wang, Ridhima Phatak; Writing - original draft preparation: Hong Wang, Zhao Han; Writing - review and editing: Zhao Han, Hong Wang, Ridhima Phatak, James Ocampo; Funding acquisition: Zhao Han; Resources: Zhao Han; Supervision: Zhao Han, Hong Wang.

**Funding** This work is supported by the U.S. National Science Foundation (NSF) under Award No. 2529206.

**Data, Materials, and Code Availability** All experiment materials, questionnaires, code, videos, data, and data analysis scripts are available in an Open Science Framework (OSF) repository: https://osf.io/5f79e/.

## Declarations

**Ethical Approval** This study was approved by the Institutional Review Board (IRB) at the University of South Florida.

**Consent for Publication** Participants signed informed consent to participate and for publication.

**Competing Interests** The authors have no relevant financial or non-financial interests to disclose.

**Hong Wang** is a Ph.D. student in Computer Science and Engineering at the University of South Florida. Her research interests are human-robot interaction, augmented reality, and computer vision.

**Ridhima Phatak** is a M.S. student in Computer Science at the University of South Florida. Her research interests are human-robot interaction and robot ethics.

**James Ocampo** was an undergraduate student in Computer Science at the University of South Florida. His research interests are human-robot interaction and augmented reality.

**Zhao Han** is an assistant professor in the Bellini College of Artificial Intelligence, Cybersecurity and Computing at the University of South Florida. His research interests are human-robot interaction, augmented reality, AI, and robotics.